%% file: main.tex
\documentclass[journal]{IEEEtran}
\usepackage{ragged2e}
\usepackage{amsmath,amssymb,amsfonts}
\usepackage{cite}
\usepackage{algorithm}
\usepackage{algorithmic}
\usepackage{graphicx}
\usepackage{physics}
\usepackage{caption}
\usepackage{lipsum}
\usepackage{enumerate}
\usepackage{amsthm}
\usepackage{color,soul}
\usepackage{xcolor}
\usepackage{booktabs}
\usepackage{threeparttable}
\usepackage{multirow}
\usepackage[switch]{lineno}
\usepackage[]{hyperref}
\hypersetup{bookmarksnumbered=true}
\newcommand{\Exp}{\mathop{\mathbb E}\displaylimits}

\newcommand\VECTOR{}  

\ifCLASSOPTIONcompsoc
  \usepackage[caption=false,font=normalsize,labelfont=sf,textfont=sf]{subfig}
\else
  \usepackage[caption=false,font=footnotesize]{subfig}
\fi

\begin{document}

\title{Improve Generalization of Driving Policy at Signalized Intersections with Adversarial Learning}

\author{Yangang Ren, Guojian Zhan, Liye Tang, Shengbo Eben Li*, Jianhua Jiang and~Jingliang Duan
\thanks{This work is supported by Beijing Science and Technology Plan Project with Z191100007419008. It is also partially supported by Tsinghua University-Tencent Joint Laboratory, and Tsinghua University-Toyota Joint Research Center for AI Technology of Automated Vehicle. All correspondences should be sent to S. Eben Li with email: lisb04@gmail.com.}
\thanks{Y. Ren, G. Zhan, L. Tang, and S. Eben Li are with State Key Lab of Automotive Safety and Energy, School of Vehicle and Mobility, Tsinghua University, Beijing, 100084, China. {\tt\small Email: (ryg18, zgj21, tly20)@mails.tsinghua.edu.cn; lishbo@tsinghua.edu.cn}.
}
\thanks{J. Jiang is with College of Engineering, China Agricultural University, Beijing, 100084, China. {\tt\small Email: jiangjianhua\_1998@163.com}.}
\thanks{J. Duan is with Department of Electrical and Computer Engineering, National University of Singapore, Singapore. {\tt\small Email: duanjl@nus.edu.sg}.
}
}

\maketitle
\begin{abstract}
Intersections are quite challenging among various driving scenes wherein the interaction of signal lights and distinct traffic actors poses great difficulty to learn a wise and robust driving policy. 
Current research rarely considers the diversity of intersections and stochastic behaviors of traffic participants.
For practical applications, the randomness usually leads to some devastating events, which should be the focus of autonomous driving.
This paper introduces an adversarial learning paradigm to boost the intelligence and robustness of driving policy for signalized intersections with dense traffic flow.
Firstly, we design a static path planner which is capable of generating trackable candidate paths for multiple intersections with diversified topology. Next, a constrained optimal control problem (COCP) is built based on these candidate paths wherein the bounded uncertainty of dynamic models is considered to capture the randomness of driving environment.
We propose adversarial policy gradient (APG) to solve the COCP wherein the adversarial policy is introduced to provide disturbances by seeking the most severe uncertainty while the driving policy learns to handle this situation by competition.
Finally, a comprehensive system is established to conduct
training and testing wherein the perception module is introduced and the human experience is incorporated to solve the yellow light dilemma. Experiments indicate that the trained policy can handle the signal
lights flexibly meanwhile realizing the smooth and efficient passing with a humanoid paradigm. Besides, APG enables a large-margin improvement of the resistance to the abnormal behaviors and thus ensures a high safety level for the autonomous vehicle. 
\end{abstract}


\IEEEpeerreviewmaketitle

\input{content/1Intro}

\input{content/2Preliminaries}

\input{content/3Method}
\input{content/4Simulation}

\input{content/5Result}

\input{content/6Conclusion}



\ifCLASSOPTIONcaptionsoff
  \newpage
\fi

\bibliographystyle{ieeetr}
\bibliography{reference}

\end{document}

%% file: content/1Intro.tex
\section{Introduction}
\IEEEPARstart{A}{utonomous} driving is regarded as one of the most potential applications of artificial intelligence, which can provide numerous benefits from increasing efficiency to reducing accidents.
Technically, decision-making is the core and challenging element to achieve high-level intelligence of the automated vehicles\cite{pendleton2017perception}.
To that end, reinforcement learning (RL) has been widely adopted to learn a reasonable driving policy inspired by its successful application in Go game and robotics\cite{silver2017mastering, rowland2018distributional_ana}. Generally, RL can be viewed as a trial-and-error learning by interacting with the environment\cite{kiran2021deep}.
While early RL applications were mainly deployed in the high-way scenes to learn some simple behaviors such as lane keeping\cite{sallab2016end}, lane changing\cite{wang2018reinforcement} or overtaking \cite{ngai2011multiple}, it is currently promising to conquer more complicated urban scenarios. 
Especially, intersection is the typical representative of urban roads due to the highly dynamic and stochastic characteristics caused by the interaction of signal lights and distinct traffic actors including pedestrians, cyclists and vehicles\cite{wu2013safety}.
In that situation, assuring driving safety has always been the principal consideration and challenge to the application of RL.

The intuitive approach for safety is introducing the constraints and updating the driving policy towards the direction of satisfying these constraints.
For that purpose, Isele \emph{et.al} (2018) firstly considered the safe exploration during the learning process by using the prediction as a constraint and claimed that it can be used to safely learn intersection handling behaviors on an autonomous vehicle \cite{isele2018navigating}. 
Wen \emph{et. al} (2020) considered a risk function and bounded the expected risk within predefined hard constraints, showing that this could assure the driving safety at the simple two-vehicle interacting at an intersection\cite{wen2020safe}. 
Besides, some other works also have explored different constraints formulation for driving tasks such as control barrier function\cite{ma2021model}, chance constraint\cite{Peng2020ModelBasedAW} and constraints in continuous time\cite{DUAN2022128}. Inspired by these constraint optimizations, Guan \emph{et.al} (2021) recently proposed the integrated decision and control (IDC) framework for autonomous driving which intuitively adopted RL to solve the constrained optimal control problems (COCP) based on multiple candidate paths generated by static path planner\cite{guan2021integrated}. The dynamic model of the ego vehicle and the empirical model of surrounding participants are built separately to construct the distance constraints.
Besides, both simulation and real vehicle experiments have proven that IDC shows more flexibility and intelligence at complex urban roads compared with the state-of-the-art methods\cite{jiang2021integrated}.
However, the IDC framework suffers two major issues when extending to more signalized intersections. 
One is the static path planner is hardly to generalized to diversified intersections with different shape and topology. The other is the policy trained on empirical models is too idealized to accommodate the abnormal behaviors from real driving environments.

The first issue says that the static path planner of IDC currently focuses on a certain intersection to design the candidate paths individually, whereas for other crossroads with different configuration, it should determine the reference paths one by one according to the realistic map of this intersection. 
Concretely, bizer curves \cite{chen2017accurate} is adopted to generate reference path with the guidance of control points, whose choices become the core procedure as they will determine totally the shape of reference paths.
This calls for a general approach to the automatic determination of control points which could be adaptive to the intersections of all shapes.
Also, this path generation only concerns the green light, and the following training process simulates red lights by adding virtual vehicle in front of the stop line, and treating the yellow light as the red one to force a hard deceleration at any time\cite{ren2021encoding}. 
This will lead to a rather conservative dealing with yellow light dilemma zones, where the ego vehicle must choose to stop or pass without hesitation according to traffic condition\cite{liu2012empirical}. 
To deal with signal lights, Zhou \emph{et al.} (2019) utilized the predefined rules to replan the trajectory online considering the ego vehicle's position, velocity, and the constraints of surrounding vehicles \cite{zhou2019development}.
Chen \emph{et al.} (2021) employed model predictive control (MPC) method to optimize the acceleration to drive the vehicle to pass the intersection at the current or the next green light phase\cite{chen2021mixed}.
Although the embedding of human experience will be helpful for handling traffic lights, this online optimization is difficult to be applied in practice because of the high computational burden, especially in complicated scenarios.

As for the second issue, IDC mainly focus on the safety guards against common situations by constraining the prior known environment models, while this kind of empirical models hardly reflects the randomness of driving tasks. This gap will bring disabled safety guarantees against the abnormal behaviors such as over-speeding or direction change contrary to the traffic rules. To improve the resistance to environmental disturbances, 
Pinto \emph{et. al} (2017) extended adversarial learning to RL fields wherein the protagonist policy tried to learn the control law by maximizing the accumulated reward while an adversary agent aimed to make the worst possible destruction by minimizing the same objective \cite{pinto2017robust}. These two policies were trained alternatively, with one being fixed whilst the other adapted.
After that, Pan \emph{et.al} (2019) explicitly used an ensemble of policy networks to model risk as the variance of value functions, and then train a risk-averse protagonist and a risk-seeking adversary\cite{pan_rararl}. They built the round way racing track in TORCS\cite{wymann2000torcs} on a self-driving vehicle controller and showed that it could handle substantially fewer catastrophic events caused by adversarial attacks.
Similarly, we demonstrated that adversarial learning could be helpful for the safe decision-making at intersections, wherein we used the protagonist policy to control the ego vehicle, the adversarial policy to control the two surrounding vehicles respectively\cite{renitsc2020}.
However, these works largely simplify the driving conditions of intersections, only considering the sparse and homogeneous traffic flows. Besides, their adversarial training process mainly focuses on the performance at worst-case situation, and lack of the guarantee of driving safety with the absence of distance constraints. 

This paper aims to learn robust and intelligent driving skills at signalized intersections, which could be scalable to diversified urban crossroads and can handle the abnormal behaviors from surroundings.
Based on the IDC framework, the contributions and novelty of this paper are summarized as follows:

\begin{enumerate}
\item A static path planner for urban intersections is developed to produce candidate paths, which is proved to be generalized to various crossroads.
Specifically, our path generation only considers static road information and consists of the route planning and the velocity planning. The former is capable of providing reference position for the automated vehicle and we design a automatic static route generation method using bizer curves where the control points can be adaptive to the size and topology of intersections.
The latter aims to deliver reference velocity to adjust the driving behaviors encountering different phases of the signal light, wherein we design two distinct modes, i.e., the stop velocity to encourage yielding or waiting behaviors at red lights and the pass velocity to inspire a quick ride at green lights. 

\item Based on the candidate paths, we construct a COCP wherein the bounded uncertainty of dynamic models is considered to capture the randomness of driving environment and the constraints are introduced to assure driving safety.
Novelly, we propose adversarial policy gradient (APG) to solve this COCP under the worst constraint, wherein the driving policy learns to realize reasonable operations and the adversarial policy aims to provide disturbances.
Formally, the minimax formulation is established based on the driving performance of tracking cost and safety constraints, with which the driving policy intends to minimize and the adversarial policy attempts to violate the safety constraints by seeking for the most severe uncertainty.
APG optimizes these two policies alternately, and the competition will provably make the driving policy obtain a greater ability to cope with external variations.


\item We establish a comprehensive driving environment wherein the typical intersections are built and the perception module is introduced to simulate the sensor characteristics. Besides, we incorporate the human knowledge into the online application of the trained networks to solve the yellow light dilemma. 
Simulations show that the trained policy can handle the switch of signal lights flexibly which is comparable to human drivers, meanwhile realizing the smooth and efficient passing at different intersections. Moreover, the generation tests are conducted to verify that APG improves the safety and robustness of the driving policy to resist abnormal behaviors of traffic participants.
\end{enumerate}

The paper is organized as follows. In Section \ref{sec.related_work}, we introduce the key notations and related works. Section \ref{sec:dpsr} describes the mathematical details of the static path planner and APG. Section \ref{sec:simulation} presents driving environment construction and the implementation for RL algorithms. Section \ref{sec:real_vehciel_experiments} verifies the effectiveness of the trained policy and Section \ref{sec:conclusion} concludes this paper.

%% file: content/2Preliminaries.tex
\section{PRELIMINARIES}
\label{sec.related_work}
This section will first introduce the basic principle of RL and the details of IDC, then the related works on adversarial training will be summarized.
\subsection{Principles of RL}
Formally, we describe the driving process as a standard RL setting wherein the automated vehicle interacts with the driving environment in a real-time manner.
At current state $s_t$, the vehicle will take action $u_t$ according to the driving policy $\pi$ and then the environment will return next state $s_{t+1}$ according to the environment model $f(s_t,u_t)$, i.e., $s_{t+1}  = f(s_t,u_t)$, and a scalar reward or utility function $l(s_t, u_t)$. 
Overall, RL seeks for optimizing the driving policy $\pi$ which maps $s_t$ to the action distribution, i.e., $u_t\sim\pi(s_t)$. To that end, value function $v^\pi$ is introduced to evaluate the performance of policies, which is concretely defined as the accumulated utility expectation under $\pi$ obtained from the state $s$, i.e., $v^\pi(s) = \mathbb{E}\{\sum_{t=0}^{T-1}l_t|s_0=s\}$.
Typically, most RL algorithms can attribute to the classic actor-critic architecture\cite{sutton2018reinforcement}. The critic is designed to update $v^\pi$ to evaluate the current policy by minimizing its distance from the target generated from interaction with environment.
Actor intends to find another better policy $\pi'$ by minimizing the calculated value function, i.e., $\pi'=\arg\min_{\pi} \mathbb{E}_{s\sim d}\{v^\pi(s)\}$, where $d$ is the state distribution determined by policy $\pi$. 
Meanwhile, the policy $\pi$ and value function $v^\pi$ are usually parameterized as $\pi_{\theta}$ and $v_w$ by neural networks (NNs) to handle continuous and high dimension tasks, in which $\theta$ and $w$ are parameters to be optimized. 
In that case, the gradient-based optimizations are usually employed to 
iteratively update these NNs to approximate the optimal policy $\pi^*$ and value function $v^*$.

\subsection{Adversarial Learning}
Adversarial learning is usually modelled as the minimax formulation, wherein there exists another adversary policy $\pi_\phi$ with $\phi$ as its parameters, to provide a competitor for the protagonist policy $\pi_\theta$.
Given the current state $s_t$, $\pi_\theta$ will take action $u_t$, $\pi_\phi$ will take action $\xi_t$, and then the next state $s_{t+1}$ will be reached. Whereas they will obtain distinct utility function: the protagonist gets a utility $l(s_t, u_t, \xi_t)$ while the adversary gets a utility $-l(s_t, u_t, \xi_t)$ at each time step. Under this scheme, the value function $v^\pi$ will be determined by these two policies together, and a zero-sum two-player game can be seen as the adversary maximizing the value function while the protagonist attempts to minimizing it:
\begin{equation}
\nonumber
\label{eq.adv_train}
\begin{aligned}
\min\limits_{\theta} \max\limits_{\phi} &\quad v^\pi(s) = \mathbb{E}\{\sum_{t=0}^{T-1}l(s_t, u_t, \xi_t)|s_0=s\}
\end{aligned}
\end{equation}
This optimizes both agents using an alternating procedure. Firstly, we learn the protagonist’s policy while holding the adversary’s policy fixed. Next, the protagonist’s policy is held constant and the adversary’s policy is learned. These two steps will be repeated until convergence.
Intuitively, the adversary is introduced to apply disturbance forces to the system and is reinforced–that is, it learns an optimal policy to thwart the original agent’s goal. This learning paradigm jointly trains a pair of agents, a protagonist and an adversary, where the protagonist learns to fulfil the original task goals while being robust to the disruptions generated by its adversary\cite{pinto2017robust}.


\subsection{Integrated decision and control (IDC)}
IDC serves as an effective framework of deploying RL to conquer the decision and control of autonomous vehicles, which consists of two essential modules: static path planner and dynamic optimal tracker\cite{guan2021integrated}.
The former aims to generate the candidate path set, denoted as $\Pi$, which builds all feasible paths considering static traffic information. The number of candidate paths can be set priorly with consideration of the potential target lanes and the traffic density of intersections. Dynamic optimal tracker considers tracking all candidate paths as well as assuring the safety against surrounding environment by meeting the constraints.
Given an arbitrary path $\tau \in \Pi$, its COCP involves the tracking performance, denoted as $J_{\rm track}$, and the constraints for different participants, denoted as $g(\cdot)$ to assure safety:
\begin{equation}
\label{eq.IDC_tracker}
\begin{aligned}
\min\limits_{\theta} \quad &J_{\rm track} = \mathbb{E}_{s_t \sim d}\bigg\{\sum^{T-1}_{i=0}l(s_{i|t}, u_{i|t}, \tau)\bigg\}\\
{\rm s.t.}\quad & s_{i+1|t}=f(s_{i|t}, u_{i|t})\\
&s_{0|t}=s_t, \\
&u_{i|t}\sim \pi_\theta(s_{i|t}), \\
&g(s_{i+1|t})\ge0\\
\end{aligned}
\end{equation}
where $T$ is the prediction horizon and $s_{i|t}$ is the driving state at the prediction step $i$, starting from the current time step $t$. 
We should emphasize the $s_{0|t}=s_{t}$ indicates that the initial state $s_{t}$ is sampled from the driving environment and its following transition can be derived through the simulation model $f(\cdot, \cdot)$ based on the corresponding driving action $u_{i|t}$. The stochastic driving policy $\pi_\theta(s_{i|t})$ aims to build the mapping from the driving states to the distribution of the actions, i.e., $u_{i|t}\sim \pi_\theta(s_{i|t})$, the latter will be delivered to control the automated vehicles.
$d$ denotes the distribution of $s_t$, which usually is designed as a joint distribution of reference path $\tau$, the state of the ego vehicle and surrounding participants, the road information, etc. 
$l(s_{i|t}, \pi_{\theta}(s_{i|t}), \tau)$ denotes the utility of tracking and $g(s_{i|t})$ denotes all the constraints on $s_{i+1|t}$, such as the distance to surrounding participants, road edge. 

Besides, the value function $v_w$ is trained to evaluate the tracking cost by predicting the track performance $J_{\rm track}$ of the given path $\tau \in \Pi$.
After training, the optimal counterpart $\pi^*$ and $v^*$ will be obtained and then applied online to implement decision and control functions. $v^*$ firstly tells the optimal path from $\Pi$ which has smaller tracking cost and less potential to collide with other surrounding vehicles. And then $\pi^*$ takes this path to construct the corresponding driving state and consequently outputs the control commands for the automated vehicle. 
Comparatively, IDC owns high online computing efficiency because of the fast forward propagation of NNs, which could calculate the control commands within 10ms at one complex intersection \cite{guan2021integrated,jiang2021integrated}. With the power of RL, it can also cast off the tedious human designs and improve the safety and intelligence level of decision and control.

%% file: content/3Method.tex
\section{Methodology}
\label{sec:dpsr}
This section introduces the methodology to improve the generalization ability of RL-enabled algorithms under the scheme of IDC from two aspects. One is the static path planner which is scalable to different intersections; the other is the APG algorithm, which could improve the resistance of the driving policy to the abnormal behaviors of traffic participants.

\subsection{General Static Path Planner}
\label{sec:planner}
\begin{figure}[ht]
\centering
\includegraphics[width=0.45\textwidth]{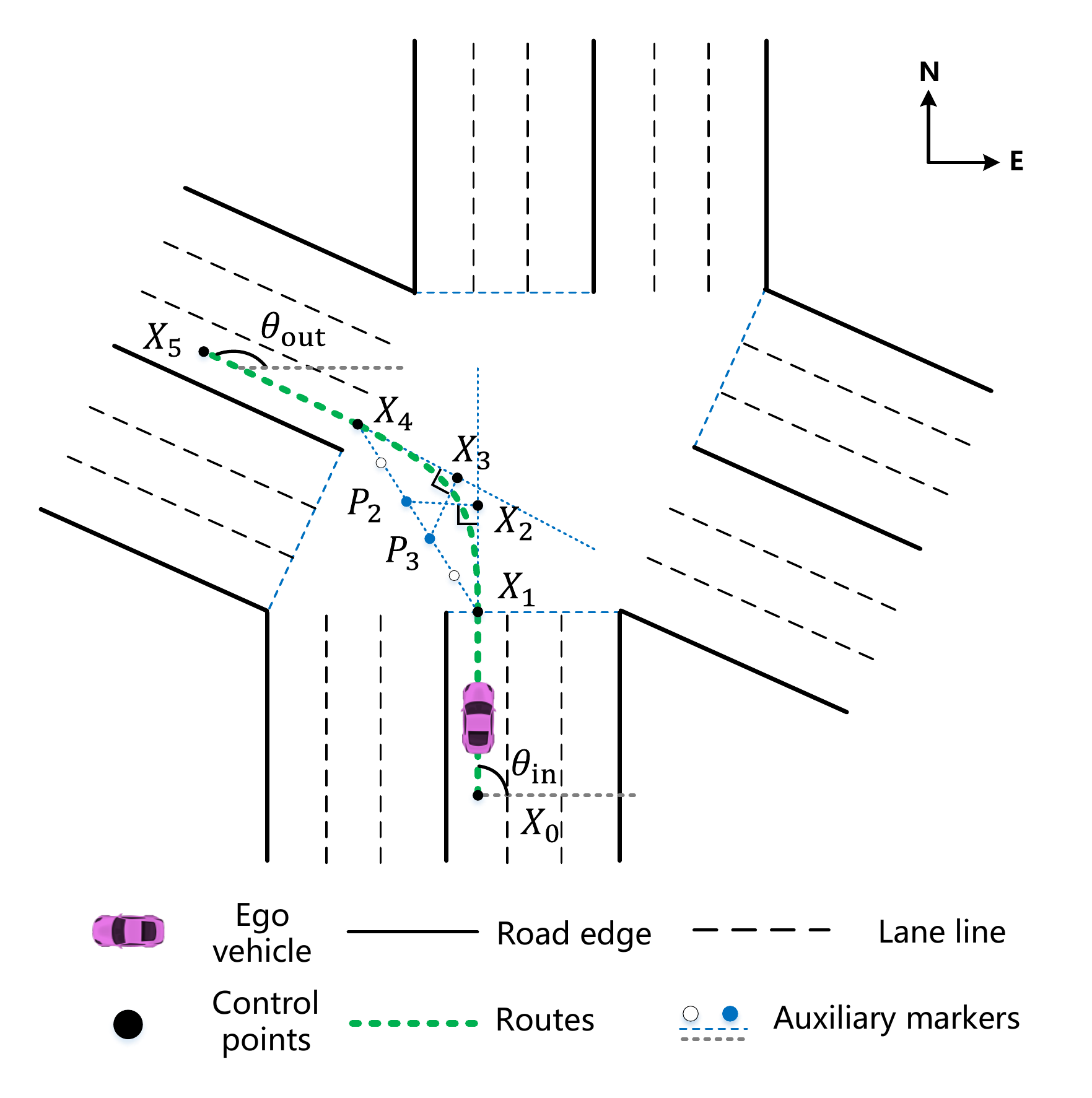}
\caption{\label{fig.route planning} Route planning for intersections}
\end{figure}

Static path planner aims to generate trackable paths with consideration of static road information. Considering the diversity and complexity of urban roads, this module should be scalable to intersections of different shapes and sizes, meanwhile incorporated with various traffic rules, such as the speed limits or pass restrictions. 
To that end, our general static path planner consists of two essential parts: route planning and velocity planning. 
For the route planning, the basic idea is to generate smooth and continuous curves by fitting and splicing, as shown in Fig.~\ref{fig.route planning}. 
Each candidate path, numbered as $\tau$, will be determined by the six control points $(X_{0}, X_{1},X_{2}, X_{3}, X_{4}, X_{5})$, which divide this candidate path into three important segments: the entrance route controlled by $(X_{0}, X_{1})$, the curve route controlled by $(X_{1}, X_{2}, X_{3}, X_{4})$ and the exit route controlled by $(X_{4}, X_{5})$.
As the routes outside intersections possess definite lanes, we can directly choose the lane center as the entrance and exit route, indicating that $(X_{0}, X_{1})$ and $(X_{4}, X_{5})$ can be constructed directly along the direction of lane center line.
For the curve route, the third-order bezier curve \cite{han2008novel} is adopted to generate a reasonable route inside intersections, which also smoothly connects the entrance route and the exit route.
Therefore, the key to generating such a curve lies in the choice of four control points $(X_{1}, X_{2}, X_{3}, X_{4})$. To guarantee the continuity of the whole route, $X_{1}$ should be the end point of the entrance lane and $X_{4}$ is the start point of the exit lane. Thus, the curve route is mainly determined by these two middle control points $(X_{2}, X_{3})$.

To assure the scalability to various intersections, we propose the choice standard of the control points $(X_{2}, X_{3})$, which can be adaptive according to the topology of the crossroads.
Without loss of generality, we assume there exists an included angle between the entrance route and the exit route with the horizontal direction, denoted as $\theta_{\rm in}$ and  $\theta_{\rm out}$ respectively, also shown in Fig.~\ref{fig.route planning}. 
The control points $(X_{2}, X_{3})$ can be decided with such a geometric method:
firstly, given $X_{1}$ and $X_{4}$, we connect them directly, thereafter two
$\rho$ bisection points, $P_{2}$ and $P_{3}$ are built, which starts from $X_{1}$ and $X_{4}$ respectively. Note that $\rho \in (0,1)$ is a hyperparameter and can be determined in experience. Then, we extend the entrance route and exit route along the inside of intersection, towards which we draw perpendicular line from $P_{2}$ and $P_{3}$ respectively. Finally, the vertical feet are determined as $X_{2}$ and $X_{3}$ respectively.
Based on the geometric relation, we can derive the specific calculation formula for $X_{i,2},X_{i,3}$ respectively:
\begin{equation}
    \begin{aligned}
    \label{eq: feature points}
        X_{2} &=  \begin{bmatrix}
        \rho\cos^2{\theta_{\rm in}} + \sin^2{\theta_{\rm in}} & (\rho-1)\sin{\theta_{\rm in}}\cos{\theta_{\rm in}} \\
        (\rho-1)\sin{\theta_{\rm in}}\cos{\theta_{\rm in}} & 
        \cos^2{\theta_{\rm in}} + \rho\sin^2{\theta_{\rm in}}  
        \end{bmatrix} X_{1} \\
        &+ \begin{bmatrix}
        (1-\rho)\cos^2{\theta_{\rm in}} & (1-\rho)\sin{\theta_{\rm in}}\cos{\theta_{\rm in}} \\
        (1-\rho)\sin{\theta_{\rm in}}\cos{\theta_{\rm in}} & 
        (1-\rho)\sin^2{\theta_{\rm in}}  
        \end{bmatrix} X_{4},
        \\
        X_{3} &=  \begin{bmatrix}
        (1-\rho)\cos^2{\theta_{\rm out}} & (1-\rho)\sin{\theta_{\rm out}}\cos{\theta_{\rm out}} \\
        (1-\rho)\sin{\theta_{\rm out}}\cos{\theta_{\rm out}} & 
        (1-\rho)\sin^2{\theta_{\rm out}}  
        \end{bmatrix} X_{1} \\
        &+ \begin{bmatrix}
        \rho\cos^2{\theta_{\rm out}} + \sin^2{\theta_{\rm out}} & (\rho-1)\sin{\theta_{\rm out}}\cos{\theta_{\rm out}} \\
        (\rho-1)\sin{\theta_{\rm out}}\cos{\theta_{\rm out}} & 
        \cos^2{\theta_{\rm out}} + \rho\sin^2{\theta_{\rm out}}  
        \end{bmatrix} X_{4}\\
    \end{aligned}
\end{equation}
After obtaining four control points $(X_{1}, X_{2},X_{3}, X_{4})$, the curve route can be generated according to the bizer formula: 
\begin{equation}
\nonumber
\begin{aligned}
\label{eq: bezier curve}
B(t)=X_{1}(1-t)^3 + 3X_{2}t(1-t)^2
&+3X_{3}t^2(1-t)\\&+X_{4}t^3, t\in [0,1]
\end{aligned}
\end{equation}
where $B(t)$ is the point coordination on the bizer curve and $t$ reflects its relative position.

\begin{figure}[!htbp]
\centering
\captionsetup[subfigure]{justification=centering}
\subfloat[]{\includegraphics[width=0.48\linewidth]{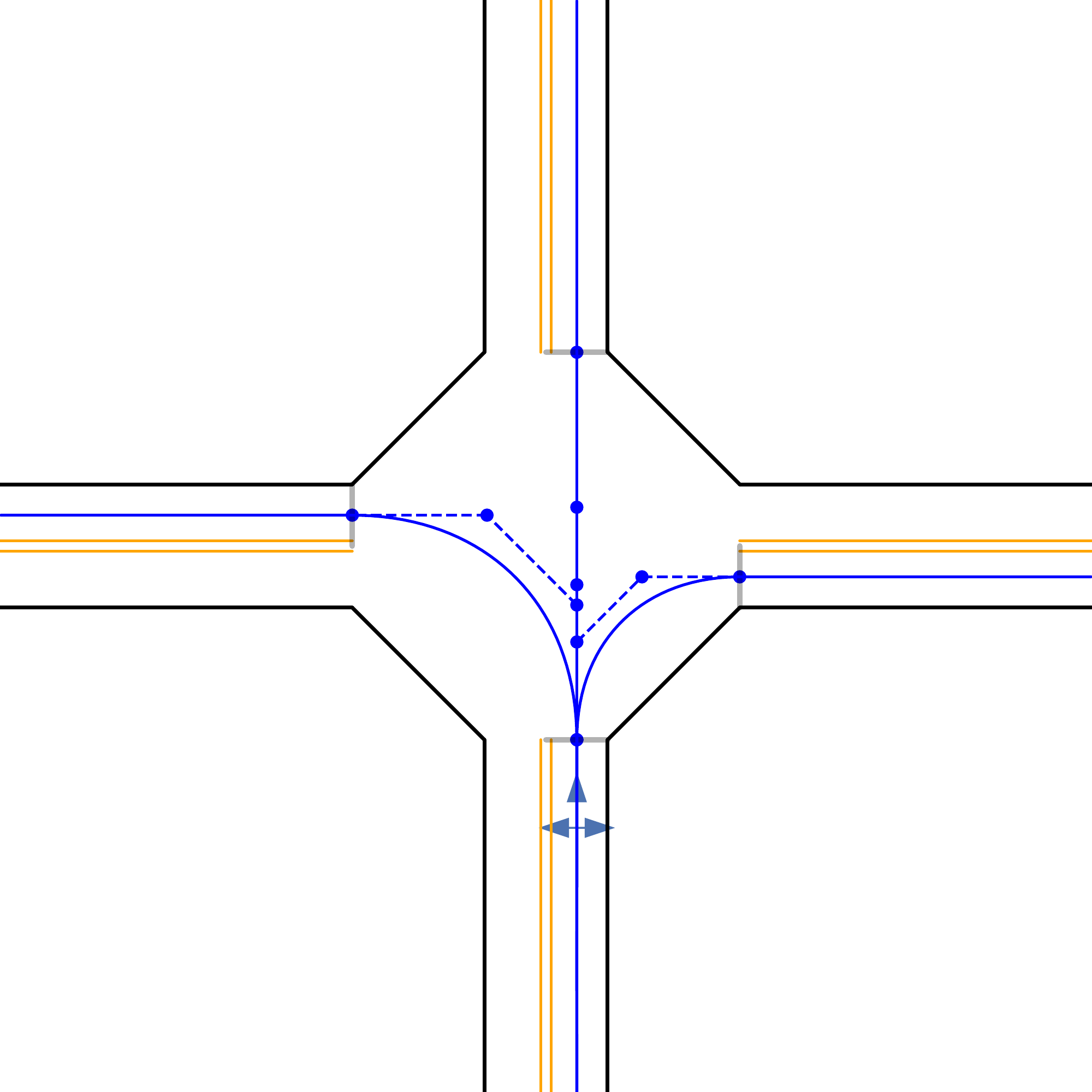}} \quad
\subfloat[]{\includegraphics[width=0.48\linewidth]{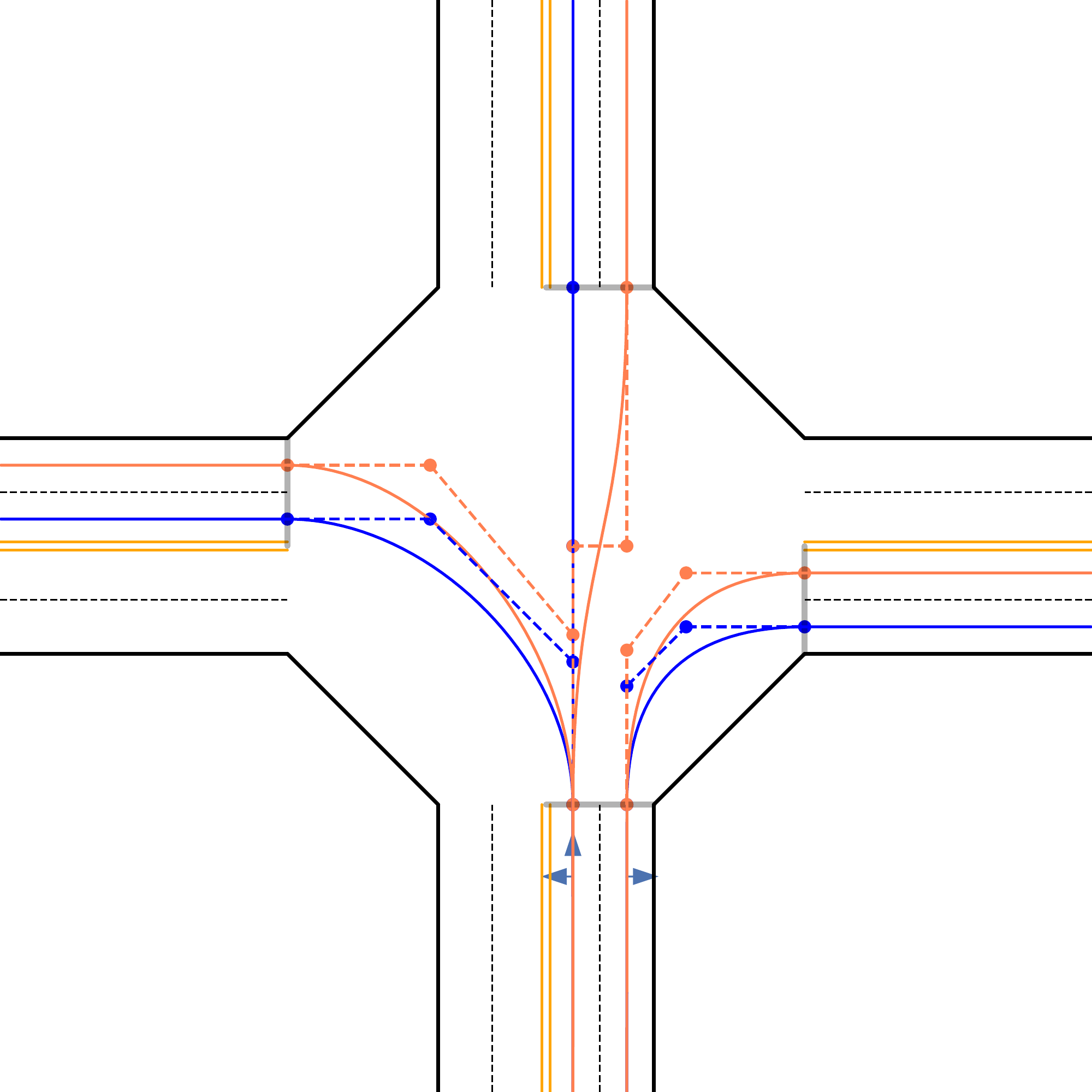}}

\subfloat[]{\includegraphics[width=0.48\linewidth]{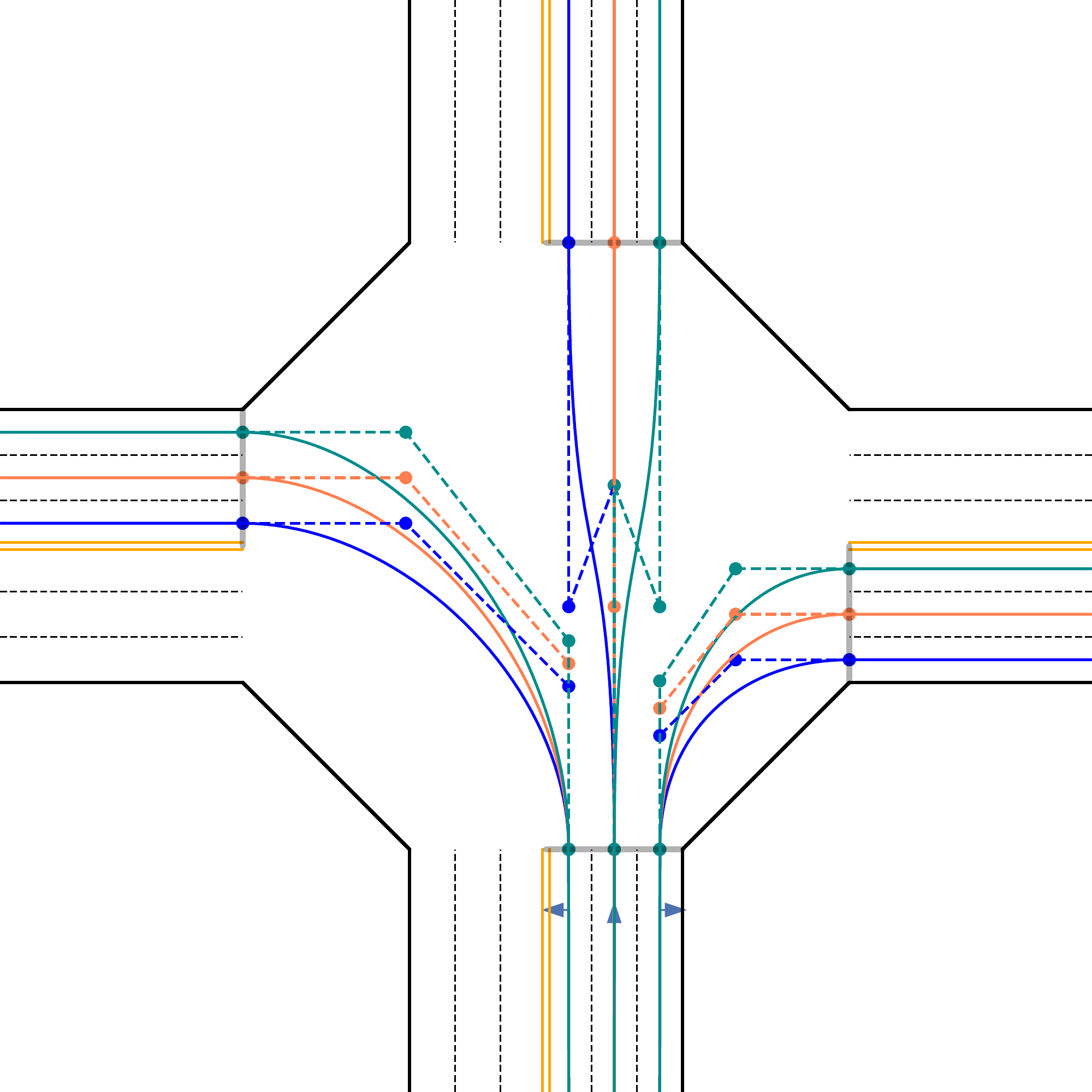}} \quad
\subfloat[]{\includegraphics[width=0.48\linewidth]{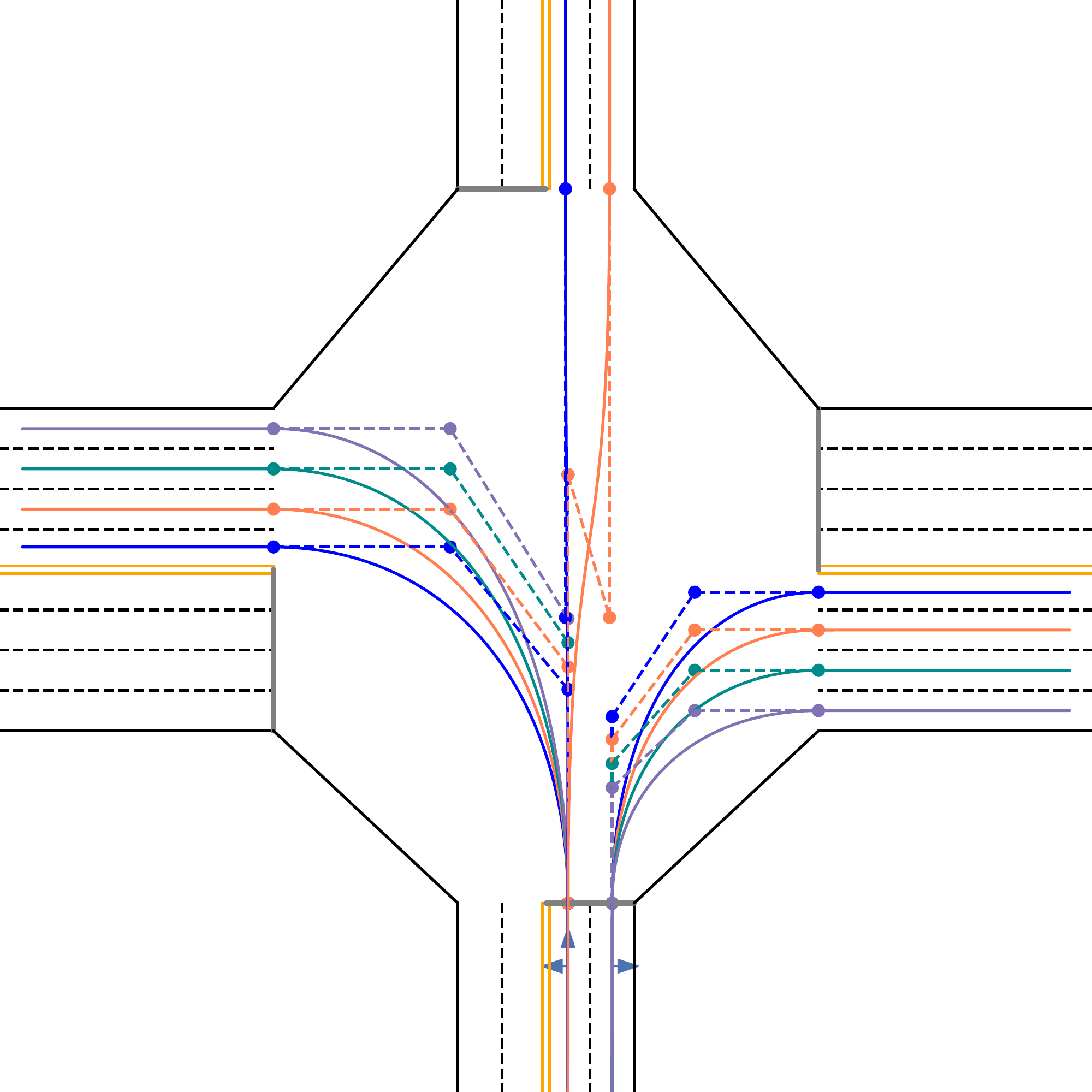}}

\subfloat[]{\includegraphics[width=0.48\linewidth]{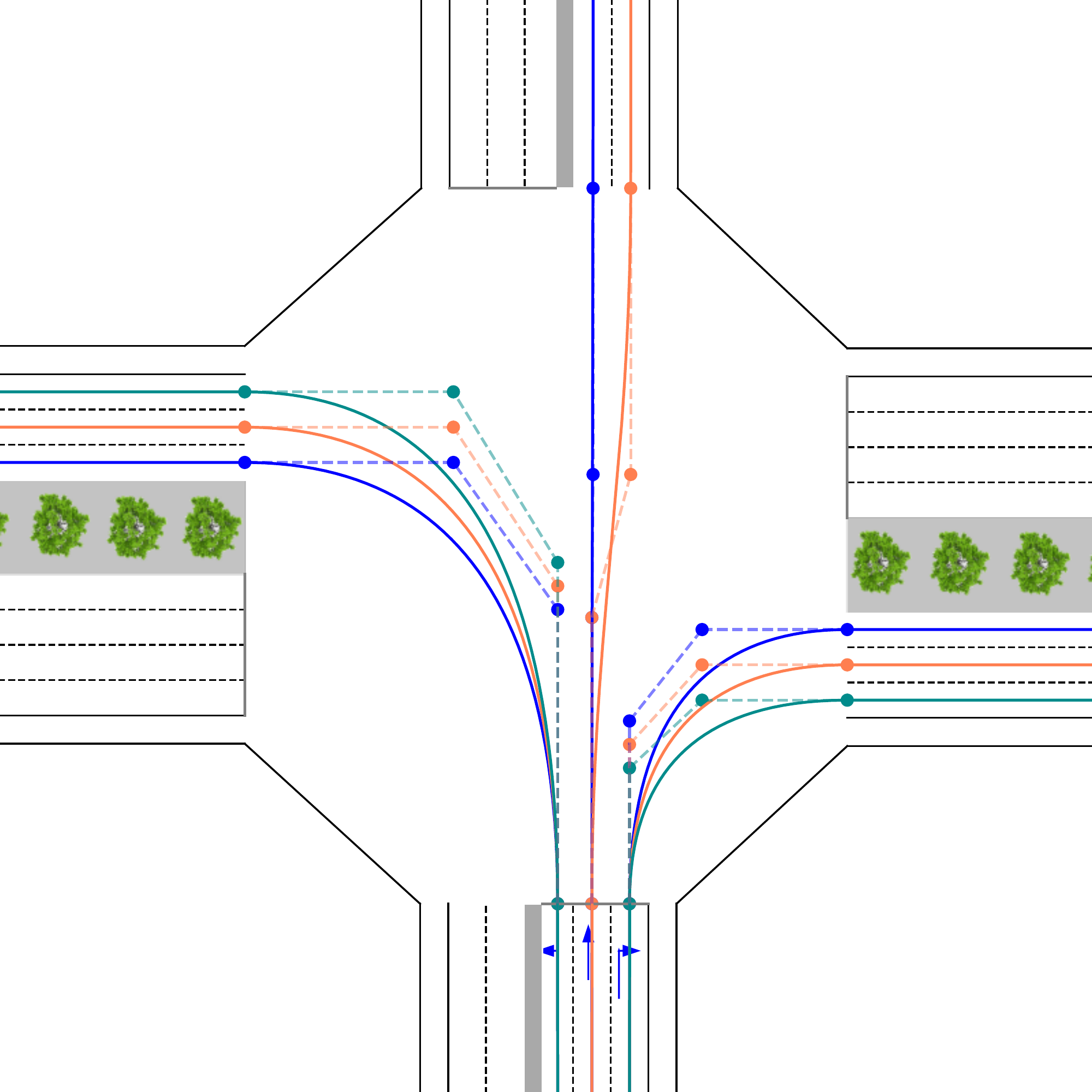}} \quad
\subfloat[]{\includegraphics[width=0.48\linewidth]{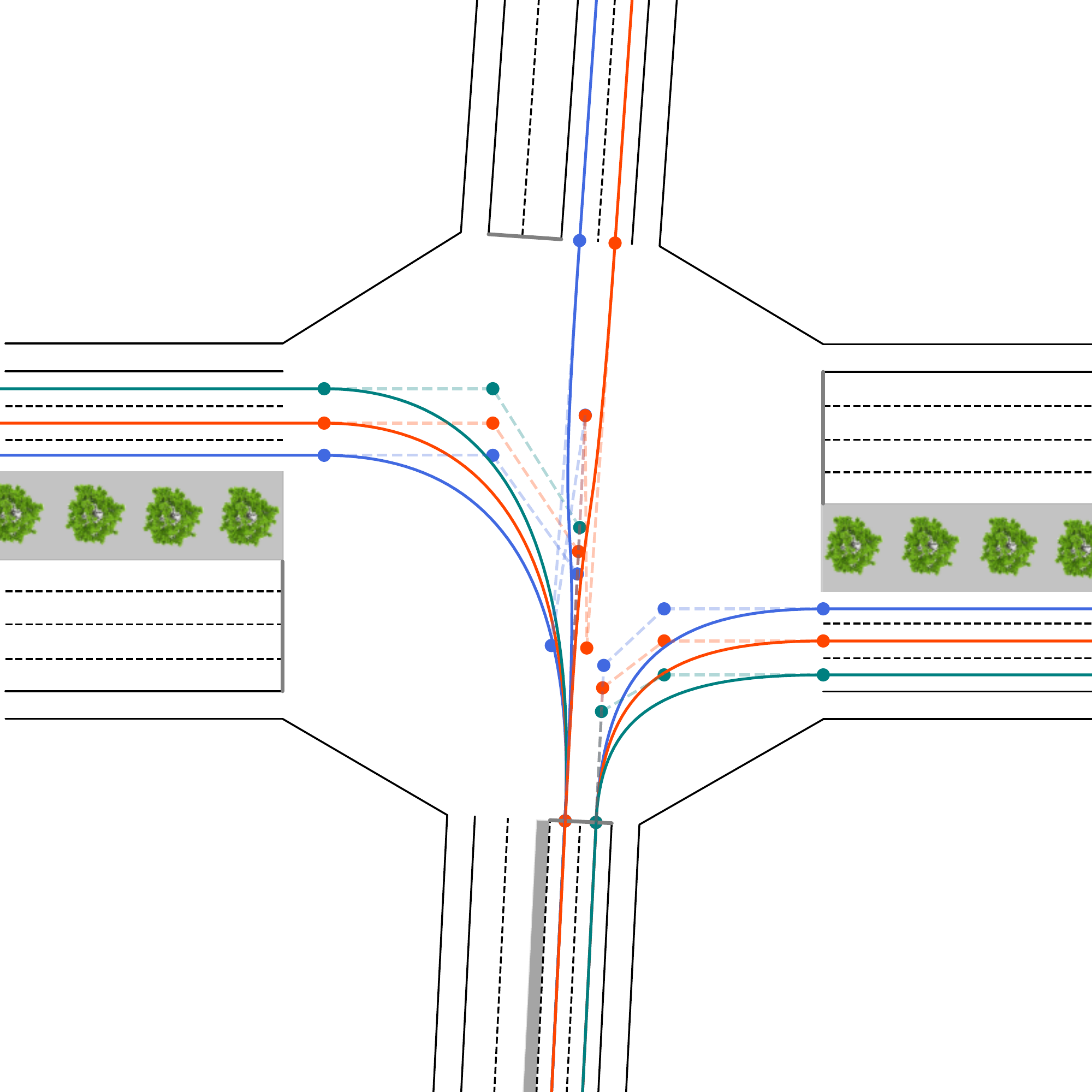}}
\caption{Route planning results at different intersections ($\rho=0.6$). The colored dots indicate the control points of bezier curves which are connected by the dotted lines. (a) Single-lane regular intersection;  (b) Two-lane regular intersection; (c) Three-lane regular intersection; (d)-(f): Multi-lane irregular intersections, where the green rectangle represents the green belts.}
\label{fig.exp_route_planning}
\end{figure}

After concatenating the entrance route, curve route and exit route together, one feasible route $\tau$ is completely generated and similarly we can construct more potential routes according to the road structure and passing connection.
To show the simplicity and high operability of this route planning, here we choose 6 typical intersections to generate the candidate routes, as shown in Fig.~\ref{fig.exp_route_planning}. These intersections are picked carefully from the open source street map to represent common scenarios, covering the single lane and multi-lane intersections, regular and irregular intersections, as well as the intersections equipped with green belts.
We suppose the ego vehicle departs from the downside and aims to finish the left-turn, right-turn and straight-going tasks. The number of candidate route is determined by that of target lane and $\rho$ is set to 0.6 for all intersections. For instance, in Fig.~\ref{fig.exp_route_planning}(f), 3 candidate routes are constructed for left-turn tasks because there exit 3 target lanes to pass through the intersection, while the straight-going task only has 2 candidate routes. 
Overall, Fig.~\ref{fig.exp_route_planning} indicates that the general static path planner can generate smooth and seemingly reasonable paths for diversified intersections with different topology and we will further verify the trackability of them in \ref{sec:V-two}.

\begin{figure}[!htbp]
\centering
\captionsetup[subfigure]{justification=centering}
\subfloat[pass velocity]{\includegraphics[width=0.4\textwidth]{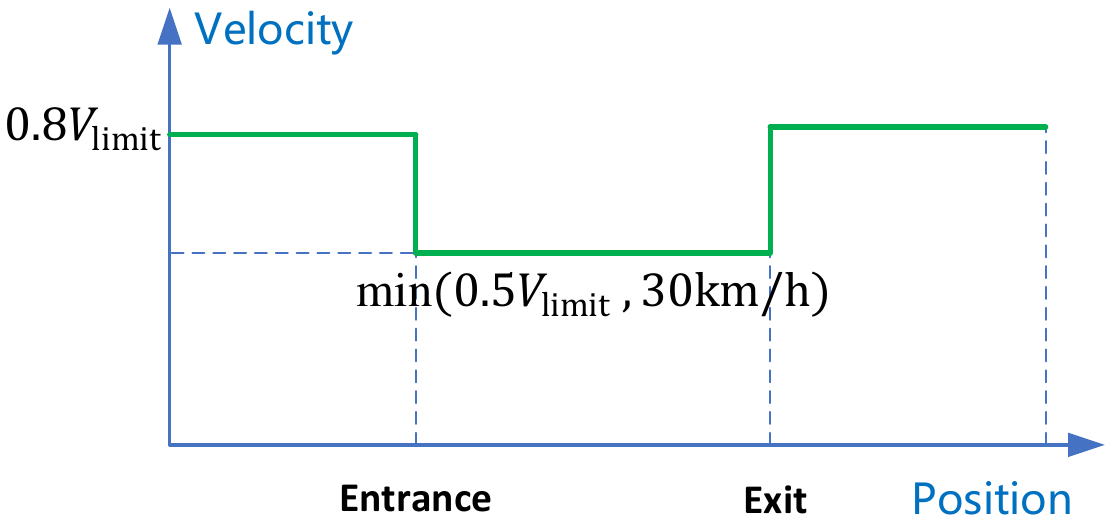}}

\subfloat[stop velocity]{\includegraphics[width=0.4\textwidth]{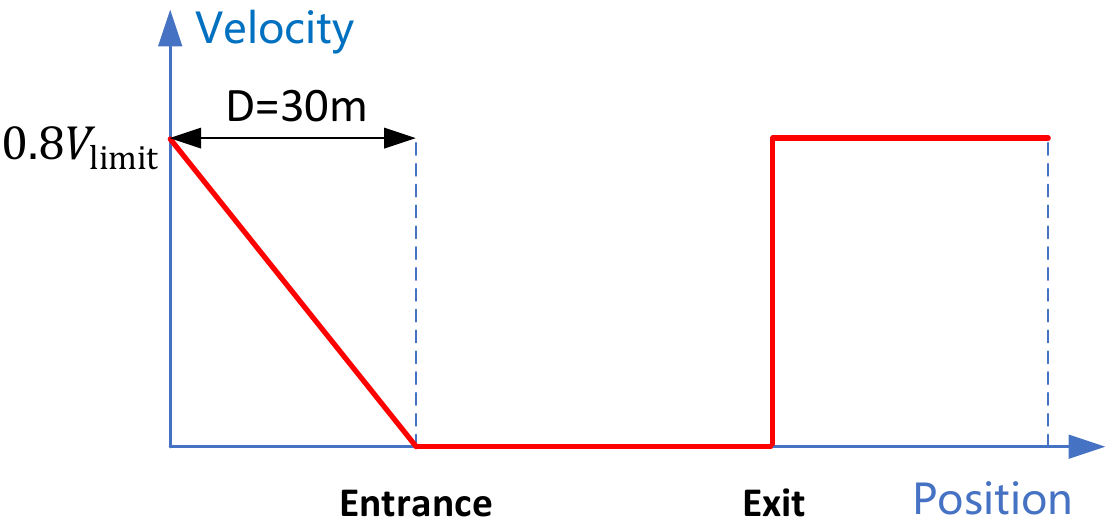}}
\caption{Velocity planning for intersections}
\label{fig.exp_velocity_planning}
\end{figure}

Another essential part of static path planner is the velocity planning, as shown in Fig.~\ref{fig.exp_velocity_planning}.
Considering the existence of traffic lights and rules at intersections, we design two expected velocity profiles with respect to the spatial position, i.e., the pass velocity mode and the stop velocity mode.
The former corresponds to passing the intersections quickly while the latter means to wait if encountering a red light or congestion inside the intersection.
For the pass velocity, its value outside the intersection is set to be $0.8V_{\rm limit}$, where $V_{\rm limit}$ shows the speed limit of the entrance lane, usually determined by traffic rules or the human experience. 
The latter is set to be the lower one between $0.5V_{\rm limited}$ and $30{\rm km/h}$ as Chinese traffic regulations usually limit the speed inside intersections to be less than $30{\rm km/h}$.
For the stop velocity, the ego vehicle is required to decelerate uniformly to the stop line within $30m$ from the speed of $0.8V_{\rm limited}$ to $0$, then the expected velocity will keep $0$ inside the intersection and turn to $0.8V_{\rm limit}$ again outside the intersection. 
Note that here we design these two velocity modes mainly considering the responses of the automated vehicle to green and red lights and both of them will be utilized to train the driving policy. After that, we further incorporate human knowledge in \ref{sec:light_deal} to apply the driving policy to deal with some unusual situations such as yellow lights. In that case, the appropriate velocity mode will be chosen in a real-time manner based on the the vehicle state and the remaining time of current light phase. 

Combining the route planning and velocity planning, the general static path planner is capable of generating candidate paths to provide the reference position and speed for automated vehicles. Note that our method only depends on the high-precision map and has no regard of the dynamic traffic participants in path production, and the subsequent training will further consider to track these paths while satisfying the safe constraints against the surrounding traffic actors.

\subsection{Adversarial Policy Gradient (APG)}
\label{sec:offline_train}
As for the construction of COCPs, it always has assumed a precise environment model, i.e., $s_{i+1|t}=f(s_{i|t}, u_{i|t})$, where the current state $s_{i|t}$ and action $u_{i|t}$ will definitely lead to a deterministic next state $s_{i+1|t}$. However, this is almost non-trivial for driving tasks due to the behavior randomness of surrounding participants and the perception noises.
Therefore, here we construct a stochastic dynamic model to consider the uncertainty of driving environment, i.e., $s_{i+1|t}=f(s_{i|t}, u_{i|t}, \xi_{i|t})$, and $\xi_{i|t}$ is the bounded random noise satisfying $\xi_{i|t} \in [\xi_{\rm min}, \xi_{\rm max}]$, where $\xi_{\rm min}$ and $\xi_{\rm max}$ represents the lower and upper bound of the randomness respectively.
Consequently, $s_{i+1|t}$ shall be a random variable and its corresponding constraint $g(s_{i+1|t})\le0$ must be held over the randomness of environment model to assure the driving safety. 
Nevertheless, this might be unavailing for the optimization process as $\xi_{i|t}$ can be captured as any value from the continuous interval $\big[\xi_{\rm min}, \xi_{\rm max}\big]$.
Intuitively, we aim to seek for the "worst case" of this safety constraint to guarantee it can work in the whole interval of the random noise, i.e., $\max_{\xi_{i|t}} g(s_{i+1|t})\le0$, which means that the most aggressive constraint w.r.t. $\xi_{i|t}$ will be satisfied and thus any other value of the random noise would be covered automatically.
To sum up, the problem formulation of our COCP can be formulized as:
\begin{equation}\label{eq.rl_problem_policy}
\begin{aligned}
\min\limits_{\theta} \quad &J_{\rm track} = \mathbb{E}_{s_t \sim d}\bigg\{\sum^{T-1}_{i=0}l(s_{i|t}, u_{i|t}, \tau)\bigg\}\\
{\rm s.t.}\quad & s_{i+1|t}=f(s_{i|t}, u_{i|t}, \xi_{i|t}),\\
&u_{i|t} \sim \pi_\theta(s_{i|t}), \\
&\xi_{i|t} \in [\xi_{\rm min}, \xi_{\rm max}], \\
&\max_{\xi_{i|t}} g(s_{i+1|t})\le0.\\
\end{aligned}
\end{equation}

This constrained optimal problem needs to be further transformed so that it can be solved by RL algorithms. Here, we adopt the generalized exterior point method \cite{guan_itsc} to convert it as the unconstrained one, wherein $\varphi(s_{i+1|t})$ is the penalty function of the state constraint $g(s_{i+1|t})\le0$, i.e.,
\begin{equation}
\nonumber
\label{eq.pely_func}
\varphi(s_{i+1|t}) = \min\{0, -g(s_{i+1|t})\}^2.
\end{equation}
Obviously, $\varphi(\cdot)$ characterizes the degree to which the constraints are satisfied, in which if $g(s_{i+1|t})\le0$ holds, the penalty will exactly be 0 otherwise it will become a square growth to the constraint violation. Based on that, we can reformulate \eqref{eq.rl_problem_policy} and construct the total policy cost $J_\pi$ as follows:
\begin{equation}
\label{eq.unconstrained_rl}
\begin{aligned}
\min\limits_{\theta}& \quad J_\pi=J_{\rm track} + \rho J_{\rm safe} \qquad \qquad \qquad \qquad  \qquad \qquad \qquad \qquad \qquad  \qquad \qquad \qquad \qquad  \qquad \qquad\\
=\Exp_{\substack{s_t \sim d\\
}} &\bigg\{\sum^{T-1}_{i=0}l(s_{i|t}, u_{i|t}, \tau)\bigg\}+\rho\Exp_{\substack{s_t \sim d\\
}}\bigg\{\sum^{T-1}_{i=0}\max_{\xi_{i|t}}\varphi(s_{i+1|t})\bigg\}\\
{\rm s.t.}\quad & s_{i+1|t}=f(s_{i|t}, u_{i|t}, \xi_{i|t})\\
&u_{i|t}\sim\pi_\theta(s_{i|t}), \\
&\xi_{i|t} \in [\xi_{\rm min}, \xi_{\rm max}], \\
&\varphi(s_{i+1|t}) = \min\{0, -g(s_{i+1|t})\}^2\\
\end{aligned}
\end{equation}
where $\rho$ is the penalty factor of the penalty function, which determines the importance level of the tracking cost $J_{\rm track}$ and safe cost $J_{\rm safe}$.
Differently, solving this problem in \eqref{eq.unconstrained_rl} needs to solve the inner maximization operator firstly to find a sequence of the worst noise for all the predictive states, i.e., $\{\xi_{0|t}, \xi_{1|t}, \dots, \xi_{T-1|t}\}$, which is rather costly for the non-linear and high-dimension optimization. Inspired by the adversarial training, we introduce the adversarial policy $\pi_\phi$ to map the state $s_{i|t}$ over the distribution of random noise $\xi_{i|t}$, i.e., $\xi_{i|t} \sim \pi_\phi(s_{i|t})$. Note that $\pi_\phi(\cdot)$ also needs to be trained progressively to reach its optimal counterpart, which can output the worst noise after receiving an arbitrary state. Thus, we can construct the minimax formulation problem:
\begin{equation}
\label{eq.minimax_form}
\begin{aligned}
\min\limits_{\theta} \max\limits_{\phi} &\quad J_\pi=\mathbb{E}_{s_{t}}\bigg\{\sum^{T-1}_{i=0}\left[l(s_{i|t}, u_{i|t}, \tau)+\rho\varphi(s_{i+1|t})\right]\bigg\}\\
{\rm s.t.}\quad & s_{i+1|t}=f(s_{i|t}, u_{i|t}, \xi_{i|t})\\
&u_{i|t}\sim\pi_\theta(s_{i|t}), \xi_{i|t} \sim \pi_\phi(s_{i|t}) \\
&\xi_{i|t} \in [\xi_{\rm min}, \xi_{\rm max}], \\
&\varphi(s_{i+1|t}) = \min\{0, -g(s_{i+1|t})\}^2\\
\end{aligned}
\end{equation}

Up to now, we can employ the policy gradient method to optimize the driving policy and adversarial policy simultaneously, wherein the former optimization aims to decrease both the tracking and safety cost, and the latter intends to seek for the worst noise to disturb the learning process. 
Concretely, $\pi_\theta(\cdot)$ conducts the minimization operation by gradient descent method, whose gradient can be derived as
\begin{equation}\label{eq.grad_policy}
\begin{aligned}
&\partial_{\theta}J_{\pi}=\partial_{\theta}J_{\rm track} + \rho \partial_{\theta}J_{\rm safe}\\
&=\Exp_{\substack{s_t \sim d\\
}}\bigg\{\sum^{T-1}_{i=0} \frac{\partial l(s_{i|t}, u_{i|t},\tau)}{\partial s_{i|t}} \frac{\partial s_{i|t}}{\partial \theta}+\frac{\partial l(s_{i|t}, u_{i|t},\tau)}{\partial u_{i|t}} \frac{\partial u_{i|t}}{\partial \theta} \\
&\qquad\qquad\qquad\qquad\qquad\qquad +\rho\frac{\partial \varphi(s_{i+1|t})}{\partial s_{i+1|t}} \frac{\partial s_{i+1|t}}{\partial \theta}\bigg\}.
\end{aligned}
\end{equation}
Similarly, the gradient of $\pi_\phi(\cdot)$ can be extracted as 
\begin{equation}\label{eq.grad_encoding}
\begin{aligned}
&\partial_{\phi}J_{\pi}=\partial_{\phi}J_{\rm track} + \rho \partial_{\phi}J_{\rm safe}\\
&=\Exp_{\substack{s_t \sim d}}
\bigg\{\sum^{T-1}_{i=0} \frac{\partial l(s_{i|t}, u_{i|t},\tau)}{\partial s_{i|t}} \frac{\partial s_{i|t}}{\partial \phi }
+\rho\frac{\partial \varphi(s_{i+1|t})}{\partial s_{i+1|t}}\frac{\partial s_{i+1|t}}{\partial \phi }\bigg\}.
\end{aligned}
\end{equation}

As for the training of value network $v_w(\cdot)$, its objective, denoted as $J_v$, aims to minimize the mean square error between the prediction of $v_w$ and the real value $J_{\rm track}$ calculated by the environment model $f(\cdot, \cdot, \cdot)$ given a random path $\tau$ and the initial state $s_t$:
\begin{equation}\label{eq.idc_critic}
\begin{aligned}
\min\limits_{w} \quad &J_{v} = \mathbb{E}_{s_{t}\sim d}\bigg\{\bigg(\sum^{T-1}_{i=0}l(s_{i|t}, u(s_{i|t}), \tau) - v_w(s_{t}, \tau)\bigg)^2\bigg\}\\
{\rm s.t.}\quad &s_{i+1|t}=f(s_{i|t}, u(s_{i|t}, \xi_{i|t})),\\
&u_{i|t}\sim \pi_\theta(s_{i|t}), \xi_{i|t} \sim \pi_\phi(s_{i|t}) \\
\end{aligned}
\end{equation}
And its gradient w.r.t. the network parameter $\theta$ can be calculated as
\begin{equation}\label{eq.grad_value}
\begin{aligned}
\nabla_{w}J_{v}=\Exp_{\substack{s_t \sim d\\}}
\bigg\{\bigg(&v_w(s_{t}, \tau) -\sum^{T-1}_{i=0}l(s_{i|t}, u_{i|t}, \tau)\bigg) \frac{\nabla v_w(s_{t},\tau)}{\nabla w} \bigg\}.
\end{aligned}
\end{equation}

Accordingly, APG is proposed to obtain a generalized driving policy which is capable of handling the uncertainty from environment model, and its implementation is based on \eqref{eq.grad_policy},  \eqref{eq.grad_encoding} and \eqref{eq.grad_value} to update the adversarial policy, the driving policy and the value function iteratively.
Once their optimal counterparts are acquired, the automated vehicle can firstly select out the optimal path $\tau^\star$ by comparing their path values, and based on that the driving state will be constructed and fed to the optimal policy $\pi_{\theta^\star}$ to generate the optimal control action $u_t^*$, i.e., 
\begin{equation}
    \begin{aligned}
    &\tau^\star = \arg \min_\tau{v_{w^{\star}}(s_{t}), \tau}, \tau \in \Pi\\
    &u_t^\star= \pi_{\theta^{\star}}(s(\tau^\star))\\
    \end{aligned}
    \label{eq.apply_value_policy}
\end{equation}
The details of APG are shown as Algorithm \ref{alg:RL-DP}. Note that we introduce the update interval $m$ to adjust the update frequency of adversarial policy and driving policy, which is an important technique to stabilize the training process as suggested in \cite{JinNJ20, HinzFWW21}.
\begin{algorithm}[!htb]
\caption{Adversarial Policy Gradient (APG)}
\label{alg:RL-DP}
\begin{algorithmic}
\STATE Initialize parameters $\theta $, $w$, $\phi $
\STATE Initialize learning rate $\beta_\theta, \beta_{w}, \beta_{\phi}$
\STATE Initialize penalty factor $\rho$
\STATE Initialize iterative step $k=0$
\STATE Initialize update interval $m$
\STATE Initialize buffer $\mathcal{B}\leftarrow\emptyset$
\STATE Initialize candidate path set $\Pi$
\REPEAT
\STATE // Sampling
      \STATE Randomly select a path $\tau \in \Pi$
      \FOR{each environment step}
      \STATE Receive $\VECTOR{s_t}$ from environment with the chosen path $\tau$
      \STATE Obtain action $u_t=\pi_{\theta}(s_t)$
      \STATE Apply $u_t$ in environment, returning $s_{t+1}$ and $l(\cdot, \cdot, \cdot)$
      \STATE Add the sample into buffer: $\mathcal{B}\cup\{s_t, u_t, l(\cdot, \cdot, \cdot), s_{t+1}\}$
      \STATE $t=t+1$
      \ENDFOR
\STATE
\STATE // Optimizing
\STATE Fetch a batch of states from $\mathcal{B}$, compute $J_{v}$ and $J_\pi$ by $f(\cdot, \cdot, \cdot)$, $\pi_{\phi}$ and $\pi_{\theta}$
\STATE Update the value function with 
\eqref{eq.grad_value}: \\
\qquad \qquad $w  \leftarrow w  - \beta_{w}\nabla_{w }J_{v}$
\STATE Update the driving policy with \eqref{eq.grad_policy}: \\
\qquad \qquad $\theta  \leftarrow \theta  - \beta_{\theta}\partial_{\theta} J_{\pi}$
\IF{$k \% m = 0$}
\STATE Update the adversary policy with \eqref{eq.grad_encoding}: \\
\qquad \quad $\phi  \leftarrow \phi + \beta_{\phi}\partial_{\phi}J_{\pi}$
\ENDIF
\STATE $k=k+1$
\UNTIL Convergence 
\end{algorithmic}
\end{algorithm}

%% file: content/4Simulation.tex
\section{Training Environment Construction}
\label{sec:simulation}
This section aims to construct the training environment of signalized intersections, and demonstrates the implement details of APG.

\subsection{Traffic Configuration}
The training environment is constructed based on six typical signalized intersections demonstrated in Fig. \ref{fig.exp_route_planning}.
The candidate path set $\Pi$ is generated with the proposed method in \ref{sec:planner} and the mixed traffic flow, i.e., the vehicles, cyclists and pedestrians, will be deployed on these intersections with the support of SUMO software\cite{SUMO2018}. 
Specifically, we set the density of pedestrians, cyclists and vehicles to 100, 100, 400 per hour for each entrance lane to simulate a dense traffic flow.
The pedestrians are controlled by the stripe model of SUMO, while the surrounding vehicles and cyclists are controlled by the the car-following and lane-changing models.
All these surrounding participants are randomly initialized at the beginning of each experiment and ride along the predefined destination.
Moreover, the signal light systems are also designed to regulate the passing of large-scale traffic flow, where the light controlling the right-turn keeps green, and that dominating the left-turn and straight-going keeps synchronous, which will lead to more complex driving operations for the automated vehicle, i.e., the unprotected left-turn. Empirically, the phase time of red, yellow and green are set to 40s, 3s, and 60s respectively.
Note that we only care about the red and green light during offline training, that is, the ego vehicle will attempt to track the pass velocity mode in Fig.~\ref{fig.exp_velocity_planning} if the light is green and the stop mode if it is red. The yellow lights will be further considered in the online application in \ref{sec:light_deal}.
Note that the $V_{\rm limit}$ is set to $37.5 \rm {km/h}$ for the velocity planning.
Based on the above settings, the ego vehicle is initialized outside of the intersection and aims to complete three different tasks, i.e., turning left, going straight and turning right, to pass this intersection with guaranteeing driving safety, efficiency and comfort simultaneously. 

\subsection{Perception system}
To make a more realistic training system, we equip the perception system for the ego vehicle including one camera, one lidar and three radars, whose specifications are referred to the real sensor products in market such as Mobileye camera, DELPHI ESR (middle range), and HDL-32E \cite{cao2020novel}. 
As shown in Table \ref{tab:Sensor parameters}, the effective perception ranges of the camera, radar and lidar are set as 80m, 60m and 70m respectively, and the horizontal field of view of them are set as $\pm 35^\circ$, $\pm 45^\circ$, $360^\circ$ respectively. During riding, only the surrounding participants entering into the perception range can be captured to construct the driving states, forming the interested participant set $\mathcal{I}$.
\begin{table}[!htb]
\centering
\caption{Sensor parameters}
\label{tab:Sensor parameters}
\begin{tabular}{cccc}
\toprule
Type & Reference Product  &Detect range &Detect angle \\
\midrule
Lidar & VelodyneHDL-32E & 70m& $360^o$\\
Camera & Mobileye camera & 80m& $\pm 35^o$\\
Radar & DELPHI ESR & 60m & $\pm 45^o$\\
\bottomrule
\end{tabular}
\end{table}

Besides, we model the perception noise by adding its statistical counterpart to the true value given by SUMO. These noises obey Gaussian distribution and their parameters, i.e., the mean and variance, come from statistics of real sensor data. Take the lidar as an example, we firstly collect the pairs of true value and observation based on the open-source dataset\cite{yin2021center}, the difference between which can be seen as the sensor error of lidars.
Then we draw the statistical histogram of the sensor error of different traffic participants and estimate the distribution parameters using maximum likelihood estimation. 
Table \ref{tab:Noise parameters} shows the distribution parameters of different traffic participants of lidars. For each surrounding participant, numbered $j$, i.e., $j \in \mathcal{I}$, the observation consists of six variables: relative lateral position $p_x^{j}$, relative longitudinal position $p_y^{j}$, relative speed $v^{j}$, heading angle $\varphi^{j}$, length $L^{j}$ and width $W^{j}$. We can see that almost all variables possess the mean close to zero and the main difference lies in their variance. 
For example, the variances of $v^{j}$ for vehicles, cyclists and pedestrians demonstrate the decrease in order as the speed of vehicles varies over a more wide range. Besides, $\varphi^{j}$ of pedestrians is much larger because their small sizes will cause difficulty to the shape detection of lidars.
\begin{table}[!htb]
\centering
\caption{Noise parameters of lidars (mean, variance)}
\label{tab:Noise parameters}
\begin{tabular}{llll}
\toprule
Type & Vehicle  &Cyclist &Pedestrian \\
\midrule
$p_{x}^{j}$ & $(-0.002,0.157^2)$ & $(0.001,0.172^2)$& $(0.001,0.110^2)$\\
$p_{y}^{j}$ & $(-0.001,0.151^2)$ & $(-0.008,0.158^2)$& $(-0.001,0.111^2)$\\
$v^{j}$ & $(-0.000,0.205^2)$ & $(0.005,0.176^2)$& $(-0.000,0.119^2)$\\
$\varphi^{j}$ & $(0.000,0.054^2)$ & $(-0.014,0.171^2)$& $(-0.003,0.229^2)$\\
$L^{j}$ & $(-0.011,0.300^2)$ & $(-0.003,0.165^2)$& $(-0.000,0.147^2)$\\
$W^{j}$ & $(0.023,0.142^2)$ & $(0.035,0.109^2)$& $(-0.002,0.141^2)$\\
\bottomrule
\end{tabular}
\end{table}

\subsection{State, action and utility}
As a typical RL algorithm, the training of APG mainly involves the design of state, action and utility function. State is the extraction of original observation given by sensors, which consists of the information of ego vehicle $x_{\rm ego}$, tracking error $x_{\rm track}$ and surrounding participants $x_{\rm other}$, i.e., $s=[x_{\rm ego}, x_{\rm track}, x_{\rm other} ]^\top$. 
Concretely, $x_{\rm ego}$ is represented with a dynamic model which contains longitudinal position $p_{x}$ and lateral position $p_{y}$, longitudinal speed $v_x$ and lateral speed $v_y$, heading angle $\varphi$, yaw rate $\omega$, front wheel angle $\delta$ and acceleration $a$, i.e., $\VECTOR{x_{\rm ego}}=[p_{x},p_{y},v_x,v_y,\varphi,\omega,\delta,a]^\top$.
$x_{\rm track}$ is constructed based on $x_{\rm ego}$ and the given reference path $\tau$, including longitudinal distance error $\Delta{x}$, lateral distance error $\Delta{y}$, speed error $\Delta{v}$ and heading angle error $\Delta{\varphi}$, i.e., 
\begin{equation}
\VECTOR{x_{\rm track}}=[\Delta{x}, \Delta{y}, \Delta{v}, \Delta{\varphi}]^\top.
\end{equation}
As for $x_{\rm other}$, we will sort different traffic actors by their distance to ego vehicle based on the observation from the perception system. Then the closest 8 vehicles, 4 cyclists and 4 pedestrians of $\mathcal{I}$ will be chosen to construct $x_{\rm other}$. For each item, the description information, as shown in Table \ref{tab:Noise parameters}, can be summarized as $\VECTOR{x_{\rm other}}=[p_{x}^{j},p_{y}^{j},v^{j},\varphi^{j},L^{j},W^{j}]^\top_{\rm j \in \mathcal{I}}$.

To realize smooth lateral and longitudinal control, we utilize the derivatives of $\delta$ and $a$ as actions, i.e., $u=[\Delta\delta, \Delta a]^\top$. Considering the actuator saturation in reality, the action shall be limited to a certain range wherein we assume $\Delta\delta\in[-0.4,0.4]$ rad/s, $\Delta a\in[-4.5,4.5]$ m/${\rm{s}}^3$. Accordingly, $\delta$ and $a$ are also limited into a reasonable range to prevent some unreasonable driving operations, i.e., $\delta\in[-0.4,0.4]$ rad and $a\in[-3.0, 1.5]$ m/${\rm{s}}^2$. 
The utility $l(\cdot,\cdot, \cdot)$ is mainly related to the precision, stability and energy-saving performance of path tracking and thus we construct a classic quadratic formulation
\begin{equation}
\label{eq:track_error}
\nonumber
\begin{aligned}
l(\cdot,\cdot, \cdot)&=0.03{\Delta{v}}^2+0.8{\Delta{x}}^2+0.8{\Delta{y}}^2+30{\Delta{\varphi}}^2\\
&+0.02{\omega}^2
+5{\delta}^2+0.05{a}^2
+0.4{\Delta\delta}^2+0.1{\Delta a}^2 .
\end{aligned}
\end{equation}
For safety constraint design, we adopt the same method with \cite{ren2021encoding}, where they utilize two circles to represent the cyclists and vehicles, one circle to represent pedestrian, and the constraints between the ego vehicle and surrounding participants can be constructed by comparing the distance between their circle centers and the given safety threshold. 

\subsection{Environment Model and Uncertainty}
\label{env_model}
Environment model $f(\cdot, \cdot, \cdot)$  is established to predict the transition of driving environment within the predictive horizon in \eqref{eq.rl_problem_policy}, which typically consists of the model of ego vehicle $f_{\rm ego}$ and that of the surrounding participants $f_{\rm other}$.
The former controls the motion of the automated vehicle, and thus the precise dynamic model has been developed and verified completely \cite{ge2021numerically}, wherein the state space equation can be written as
\begin{equation}
\begin{aligned}
\nonumber
f_{\rm ego}=
\begin{bmatrix}
p_{\rm x} +\Delta t (v_{\rm x}\cos{\varphi}-v_{\rm y}\sin{\varphi})\\
p_{\rm y} +\Delta t (v_{\rm x}\sin{\varphi}+v_{\rm y}\cos{\varphi})\\
v_{\rm x}+\Delta t (a+v_{\rm y}\omega)\\
\frac{mv_{\rm x}v_{\rm y}+\Delta t[(L_f k_f-L_r k_r)\omega-k_f \delta v_{\rm x}-mv_{\rm x}^2\omega]}{mv_{\rm x}-\Delta t (k_f + k_r)}\\
\varphi + \Delta t \omega\\
\frac{-I_z \omega v_{\rm x}-\Delta t[(L_f k_f-L_r k_r)v_{\rm y}-L_f k_f \delta v_{\rm x}]}{\Delta t(L^2_f k_f+L^2_r k_r)-I_z v_{\rm x}}\\
\delta + \Delta t \Delta\delta \\
a + \Delta t \Delta a \\
\end{bmatrix}.\\
\end{aligned}
\end{equation}
Specially, the key parameters of the ego are well-designed in accordance with a physical vehicle, as shown in TABLE \ref{tab.vehicle_parameters}.
\begin{table}[tbhp]
\centering
\caption{Parameters for $f_{\rm ego}$}
\label{tab.vehicle_parameters}
\begin{tabular}{clr}
\toprule
Parameter & Meaning & Value \\
\midrule
$k_f$ & Front wheel cornering stiffness & -155495 [N/rad]  \\
$k_r$ & Rear wheel cornering stiffness & -155495 [N/rad] \\
$L_f$ & Distance from CG to front axle & 1.19 [m] \\
$L_r$ & Distance from CG to rear axle & 1.46 [m] \\
$m$ & Mass & 1520 [kg] \\
$I_z$ & Polar moment of inertia at CG & 2642 [kg$\cdot\mathrm{m}^2$] \\
$\Delta t$& Discrete time & 0.1 [s]\\ 
\bottomrule
\end{tabular}
\end{table}

$f_{\rm other}$ takes charge of predicting the motion of surrounding traffic participants, which involves the calculation of safety requirements. Usually, the kinematics models are adopted as its simplicity and good performance on structured roads\cite{hou2019interactive}.
However, this may be unable to describe the randomness of intersections, so we add the additional uncertainty terms to $\xi_x, \xi_y, \xi_v, \xi_\varphi$ to the corresponding states of this basic kinematics model and construct the following stochastic prediction model:
\begin{equation}
\label{eq:surr_model}
f_{\rm other}=
\begin{bmatrix}
p_{\rm x}^{j} +\Delta t v^j\cos{\varphi^j} + \xi_x\\
p_{\rm y}^{j} +\Delta t v^j \sin{\varphi^j} + \xi_y\\
v^j + \xi_v\\
\varphi^j + \frac{\Delta t v^j}{R} + \xi_\varphi\\
L^{j}\\
W^{j}\\
\end{bmatrix}
\end{equation}
in which $R$ indicates the turning radius determined by the route and position of the corresponding participant. And $\xi=[\xi_x, \xi_y, \xi_v, \xi_\varphi]\top$ will be generated by the adversarial policy $\pi_\phi$, i.e., $\xi \sim \pi_\phi$, to learn the aggressive noise to violate the safety constraints.

\subsection{Dealing with traffic lights}
\label{sec:light_deal}
\begin{figure}[!htbp]
\centering
\includegraphics[width=0.38\textwidth]{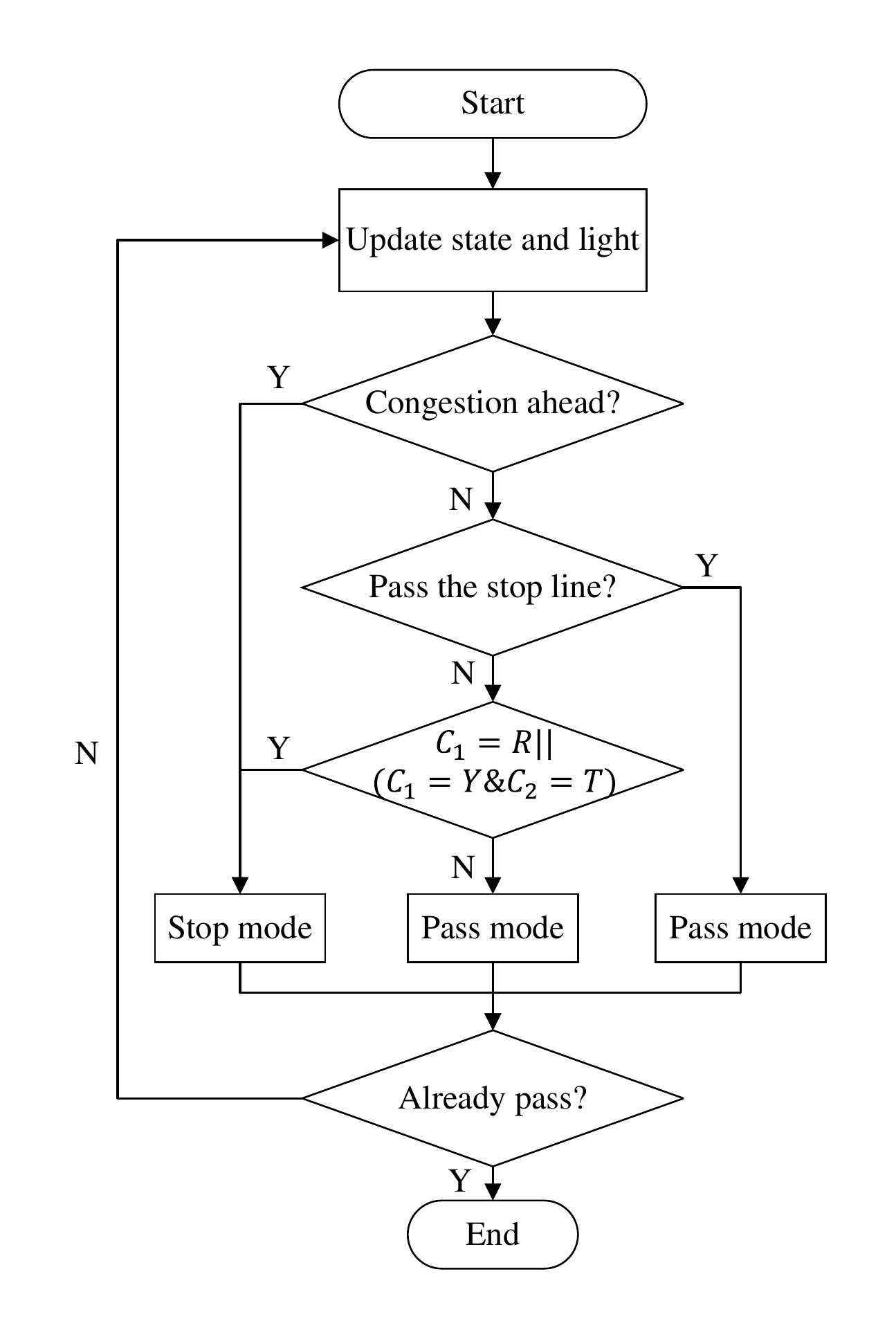}
\caption{Flowchart of dealing with signal lights.}
\label{fig.speed_mode_choice}
\end{figure}

After offline training, we have obtained $V_{w^*}$ for path selecting and $\pi_{\theta^*}$ for path tracking. As for the subsequent online application, we can further incorporate more human experience with the trained functions to handle some special traffic conditions at signalized intersections such as the yellow light or traffic congestion. 
Therefore, a series of rules are established in terms of the driving states of ego vehicle and the phase of signal lights to select the appropriate velocity mode, i.e., the pass mode or the stop mode in Fig.~\ref{fig.speed_mode_choice}.

We firstly judge whether there exists traffic congestion ahead, i.e., the front vehicle keeps stopping over 3s. When the result is yes, the stop mode in Fig.~\ref{fig.exp_velocity_planning} will be chosen to avoid further deterioration of traffic efficiency, otherwise we will judge whether the ego vehicle has passed the stop line based on the vehicle's position and the road map. If the answer is yes, the pass mode will be selected as it has entered into the intersections. Otherwise, the final judgement will be shown up to further distinguish situations, wherein the result will be jointly determined by two conditions, denoted as $C_1, C_2$ respectively: 
\begin{enumerate}
\item $C_1 \in \{R,G,Y\}$ denotes the current light phase, where $R, G, Y$ indicate the red, green and yellow lights respectively. 
\item $C_2 \in \{T, F\}$ denotes whether the ego can stop in front of the stop line before the red light comes when the current light is yellow. Note that $T, F$ are the abbreviations of TRUE and FALSE respectively.
\end{enumerate}
Therefore, the ego vehicle will choose the stop mode undoubtedly when encountering a red light, or there is enough time and distance to decelerate at the yellow light. As for other situations, for example, when the green light or yellow light appears but it is hard to stop by braking, the ego will directly pass the intersection with the guidance of pass velocity mode.

Formally, $C_1$ can be captured directly by the vehicle sensors or the communication devices bound to the signal lights. $C_2$ can be derived based on the current speed of the vehicle $v_x$ and the constant acceleration model. At each time step during the yellow light phase, the ego vehicle is assumed to brake with the 80\% of the maximum deceleration, i.e., $a_M=-2.4 \rm m/s^2$. Based on that, the longest deceleration distance $D_e$ and time $t_e$ can be deduced by
\begin{equation}
\nonumber
    \label{eq.judge_info}
    \begin{aligned}
    D_e=\frac{v^{2}_x}{2a_M}, \quad
    t_e=\frac{v_x}{a_M}.
    \end{aligned}
\end{equation}
At this moment, the distance of the ego vehicle to the stop line $D_{y}$ can be read from high precision map. Referred to Chinese traffic rules, the duration of yellow light is always 3s and thus we can obtain the remaining time, denoted as $t_y$, by counting from the initial stamp of yellow light.
Based on these conditions, $C_2=T$ holds if $D_{\rm y} \ge D_e$ and $t_y \ge t_e$ are satisfied simultaneously, otherwise $C_2=F$ is concluded to indicate that the ego can not stop in time at this constant and then the passing velocity will be given to encourage rushing before the red lights.

Note that this choice procedure is conducted at each step of the driving process, therefore the output mode will provably be changing with the state of vehicle and light duration.
Besides, the stop or pass mode only determines the speed error $\Delta{v}$ in the track error of \eqref{eq:track_error}. After that, each candidate route will be further adopted to construct other elements $\Delta{x}, \Delta{y}, \Delta{\varphi}$ respectively and $V_{w^*}$ will select the optimal one to track.

%% file: content/5Result.tex
\section{Simulation results}
\label{sec:real_vehciel_experiments}
This section mainly tells the training parameters, visualizes the training curves and tests the generalization of the driving policy.

\subsection{Training settings}
The NNs of driving policy and adversarial policy employ similar architecture,
wherein the multi-layer perceptron (MLP) is adopted as the approximate function, containing 5 hidden layers, consisting of 256 units per layer.
Considering the stochastic property of both policies, the output of MLPs are
the mean and covariance of the Gaussian distribution where the covariance matrix is diagonal. In this case, each policy maps the input states to the mean and logarithm of standard deviation of the Gaussian distribution.
Moreover, all hidden layers take GeLU as the activation function while the output layer uses the hyperbolic tangent to bound the output into the pre-defined range as shown in Table \ref{tab.hyper}. 
For the value network, it adopts almost the same architecture with the policy, except that the activation function of the output is the ReLU  to output a positive scalar to indicate the path value.
The Adam method \cite{Diederik2015Adam} with a cosine annealing learning rate is adopted to update all networks. The predictive horizon $T$ is set as 25 and the penalty factor $\rho$ is 15.
We also include the parallel training for RL as in \cite{guan2021mixed} to accelerate the training process, which contains 10 workers to sample from driving environment, 10 buffers to store samples and 18 learners to update the networks.
Table \ref{tab.hyper} provides more detailed hyperparameters of our training.
\begin{table}[!htp] 
\centering
\caption{Training hyperparameters.}
\label{tab.hyper}
\begin{threeparttable}[ht]
\begin{tabular}{lc}
\toprule
\quad Hyperparameters & Value \\
\hline
\quad Optimizer &  Adam ($\beta_{1}=0.9, \beta_{2}=0.999$)\\
\quad Approximation function  & MLP \\
\quad Number of hidden layers & 5\\
\quad Number of hidden units & 256\\
\quad Nonlinearity of hidden layer& GELU\\
\quad Batch size & 256\\
\quad Policy learning rate & cosine annealing 1e-4 $\rightarrow$ 2e-6 \\
\quad Value learning rate & cosine annealing 3e-4 $\rightarrow$ 1e-6\\
\quad Total iteration & 200000 \\
\quad Predictive horizon (T) & 25\\
\quad Update interval ($m$) & 5\\
\quad Penalty factor ($\rho$) & 15\\
\quad Range of $\xi_x$ ($m$) & [-0.8, 0.8]\\
\quad Range of $\xi_y$ ($m$) & [-0.8, 0.8]\\
\quad Range of $\xi_v$ ($m/s$) & [-0.075, 0.225]\\
\quad Range of $\xi_\varphi$ ($\rm rad$) & [-0.025, 0.025]\\
\bottomrule
\end{tabular}
\end{threeparttable}
\end{table}

\begin{figure}[!htb]
\captionsetup[subfigure]{justification=centering}
\subfloat[Total policy cost]{\label{f:return_total_policy}\includegraphics[width=0.5\linewidth]{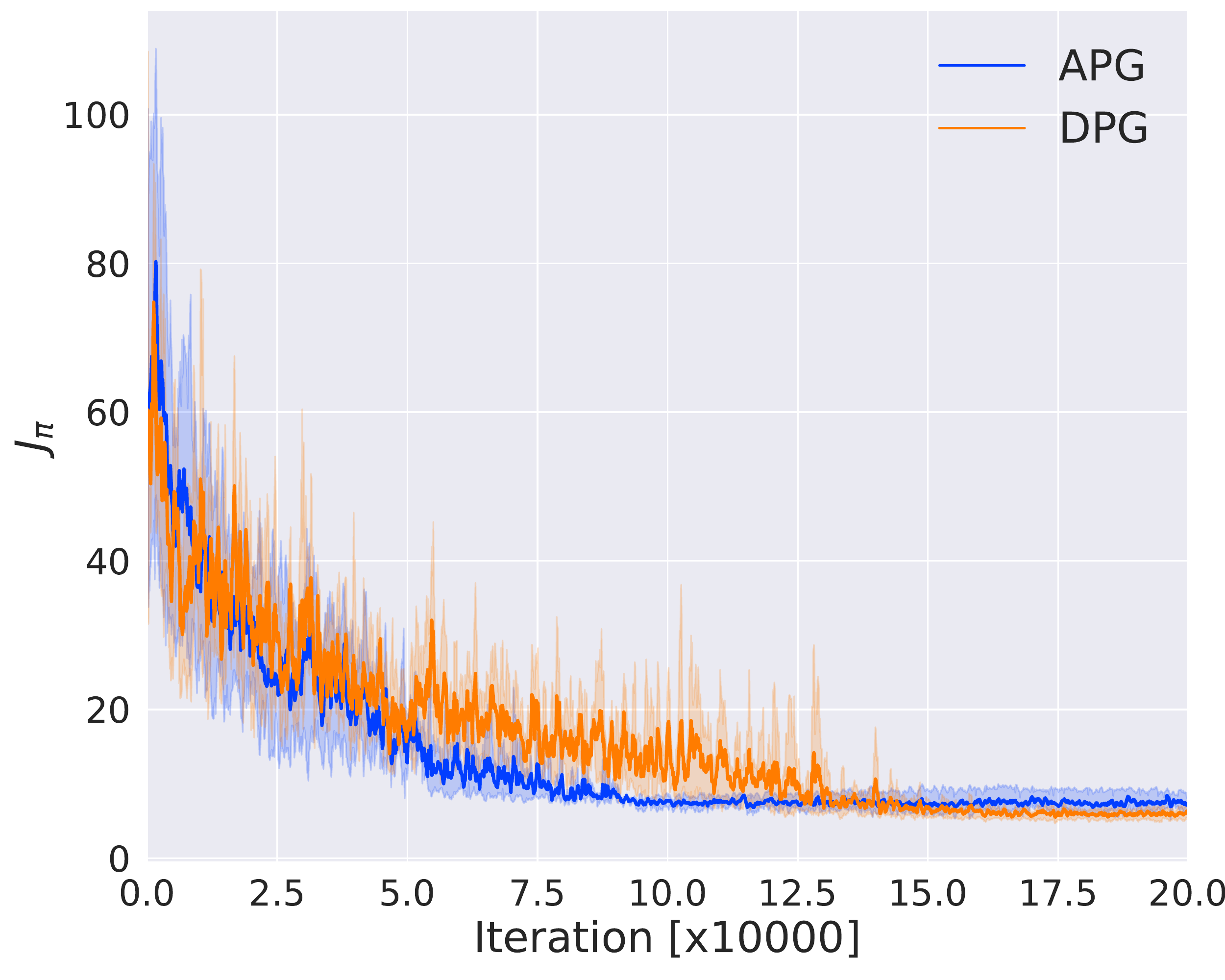}}
\subfloat[Value cost]{\label{f:return_total_value}\includegraphics[width=0.5\linewidth]{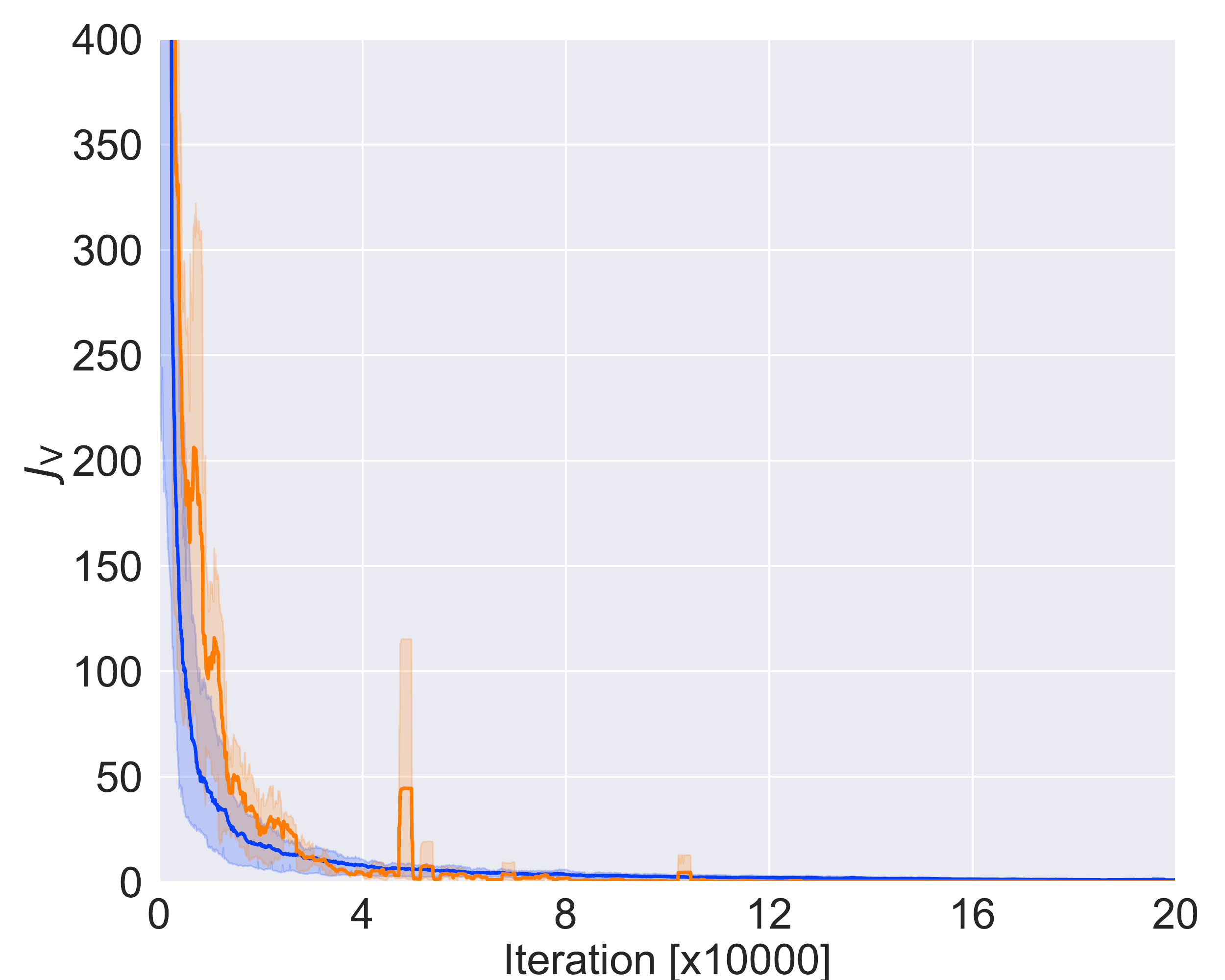}}\\
\subfloat[Tracking cost of policy]{\label{f:return_track}\includegraphics[width=0.5\linewidth]{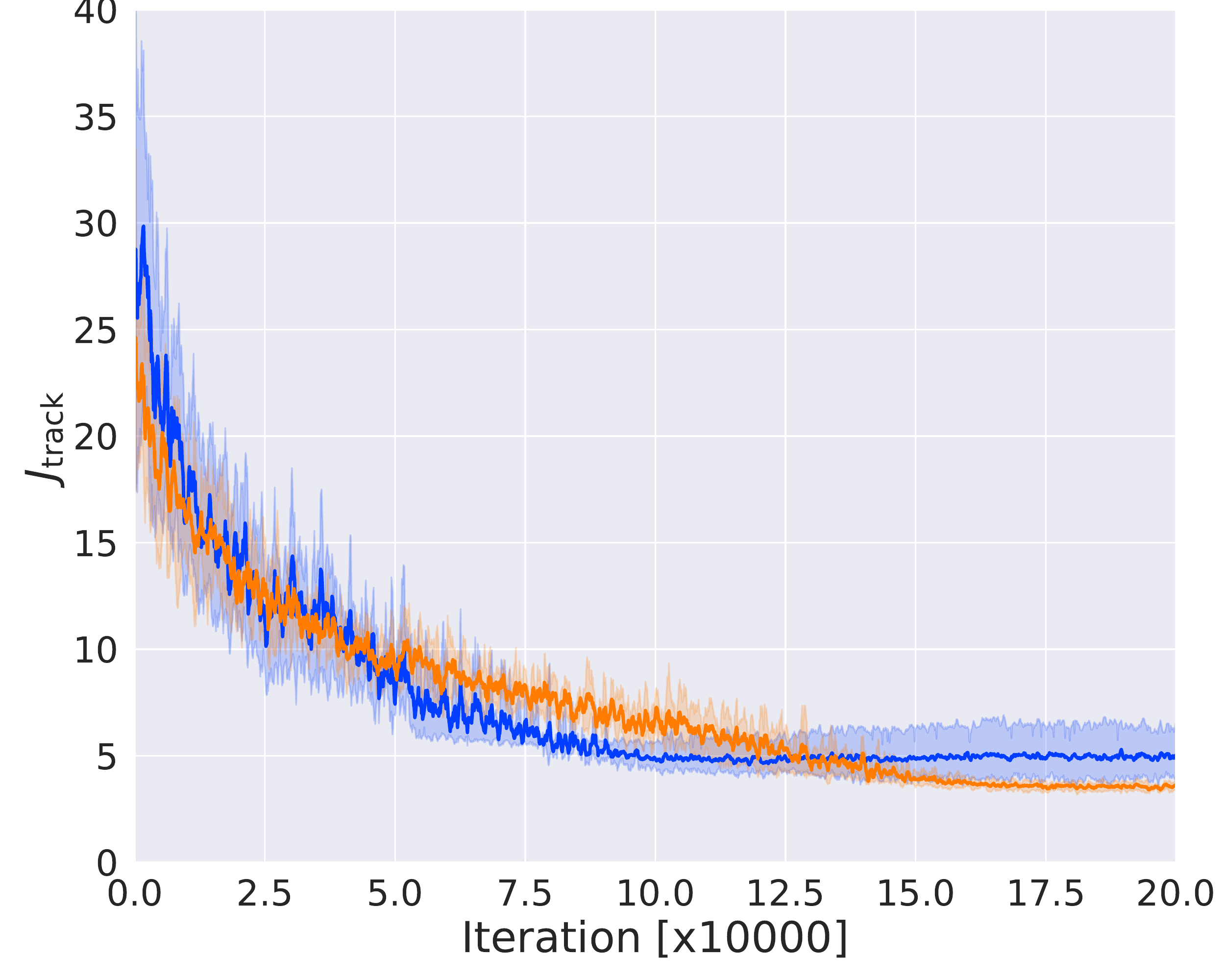}}
\subfloat[Safety cost of policy]{\label{f:return_safe}\includegraphics[width=0.5\linewidth]{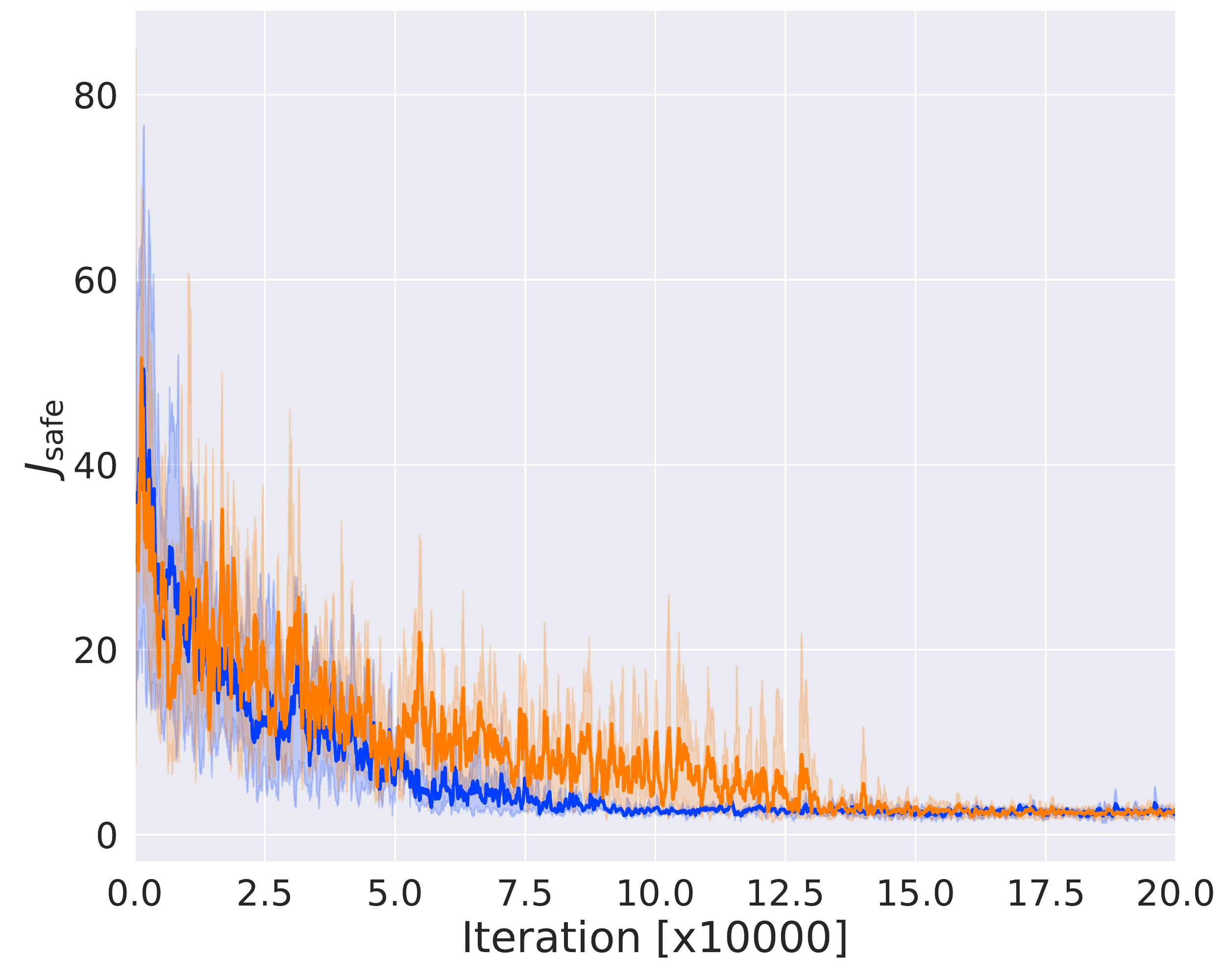}}\\
\caption{Training performance. The solid lines correspond to the mean and the shaded regions correspond to 95\% confidence interval over 5 runs.}
\label{f:return}
\end{figure}

\begin{figure}[!htbp]
\centering
\includegraphics[width=0.45\textwidth]{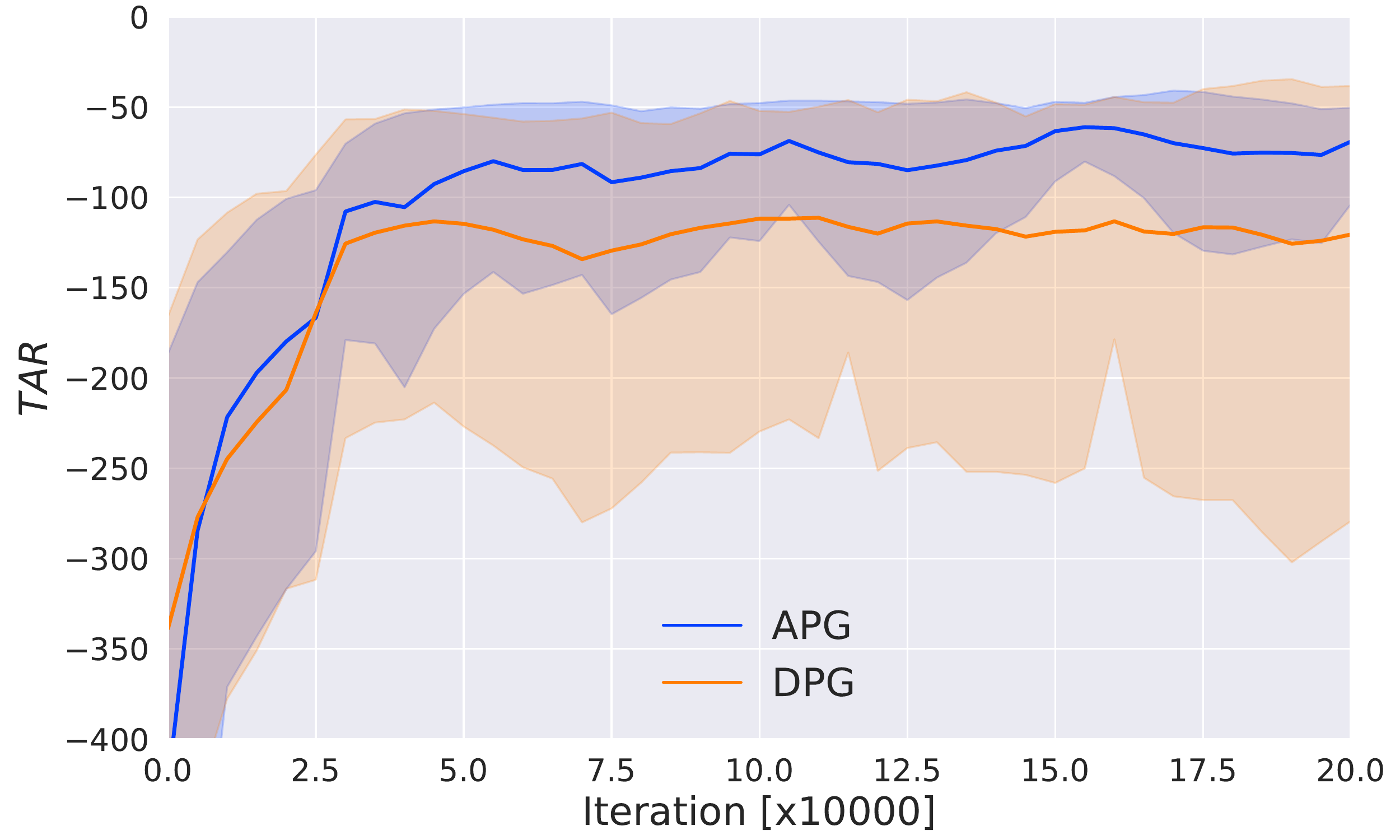}
\caption{Driving performance test on real intersections during training process. The solid lines correspond to the mean and the shaded regions correspond to 95\% confidence interval over 5 runs.}
\label{f:test-return}
\end{figure}

Then, we compare the training performance of APG with the baseline called deterministic policy gradient (DPG), wherein the only difference lies in the consideration of the randomness in the surroundings model $f_{\rm other}$.
For DPG, it assumes the empirical model $f_{\rm other}$ is accurate enough and thus the uncertainty $\xi$ is ignored to train the single driving policy.
Whereas, APG introduces the adversarial policy to simulate the despiteful noise and the driving policy must learn to handle such tough cases to make reasonable decisions. 
During the training process, we record the policy performance and value loss every 1000 iterations and visualize the learning curves in Fig.~\ref{f:return}, including the value loss $J_v$, the track loss $J_{\rm track}$, the safety loss $J_{\rm safe}$ and their summation, i.e., the total policy loss $J_{\rm \pi}$.
Note that all these performances only involve the training based on the environment model established in \ref{env_model}, not the driving performance in the real driving environment. 
To do that, we also test the temporary policies by applying them to control the automated vehicle to pass the intersection for 10 runs in the real environment every 1000 iterations.
At each run, the vehicle will be initialized outside of the intersections and move forward 120 steps during which the utility function will be accumulated. We use the total average return $(TAR)$ to reflect its real performance, i.e., 
\begin{equation}
\nonumber
TAR=\frac{1}{10}\sum^{10}_{n=1} \sum^{120}_{t=1} \left[l(s_{t}, u_{t}, \tau)+\rho\varphi(s_{t+1})\right].
\end{equation}
And Fig.~\ref{f:test-return} visualizes the $TAR$ performance during the training process.

We can see from Fig.~\ref{f:return} that all losses of these two training paradigms show a remarkable decline with the increasing of iteration, indicating the driving policy has been improved progressively with the guidance of their corresponding simulation models. 
In addition, the tracking loss and safety loss almost keep the synchronous descent until they all drop to the same level, meaning that our driving policy can balance the optimization of tracking performance and safety requirements wisely.
However, these exists a little difference between APG and DPG. The latter seems to obtain better policy performance as shown in Fig.~\ref{f:return_total_policy} because it accesses a lower track cost in Fig.~\ref{f:return_track}. 
That is, APG has to sacrifice some tracking performance to guarantee a fair safety cost, which is explicable that the adversary policy provides a strong disruption to the learning of driving policy. In this case, 
the automated vehicle has learned to avoid the potential collision by decelerating or diversion encountering the despiteful surrounding vehicles, which will lead to the decreasing of tracking performance.

As for the test performance of these two driving policies shown in Fig.~\ref{f:test-return}, we can see both of them indeed behave better and better at the real intersection, but APG has improved TAR with a large margin. 
Besides, DPG suffers a little higher fluctuations than APG shown from the confidence interval because it will be likely to fail encountering some unusual conditions.
This result is very inspiring, which says although we can realize the training convergence based on the constructed simulation model, but it might hardly work in real environment if there exists a large gap between simulation model and the real driving environment. Thus, for the model-based RL or classical control methods like Model Predictive Control (MPC), the model uncertainty should be considered especially at complex urban roads. Obviously, the kinematic model cannot describe the randomness of the intersections in our simulation, as there exists complex interactions between different traffic participants.

\subsection{Visualization}
\label{sec:V-two}
\begin{figure}[!htb]
\captionsetup{singlelinecheck = false,labelsep=period, font=small}
\centering
\captionsetup[subfigure]{justification=centering}
\subfloat[Tracking error distribution]{\label{track_error_dist}\includegraphics[width=0.5\linewidth]{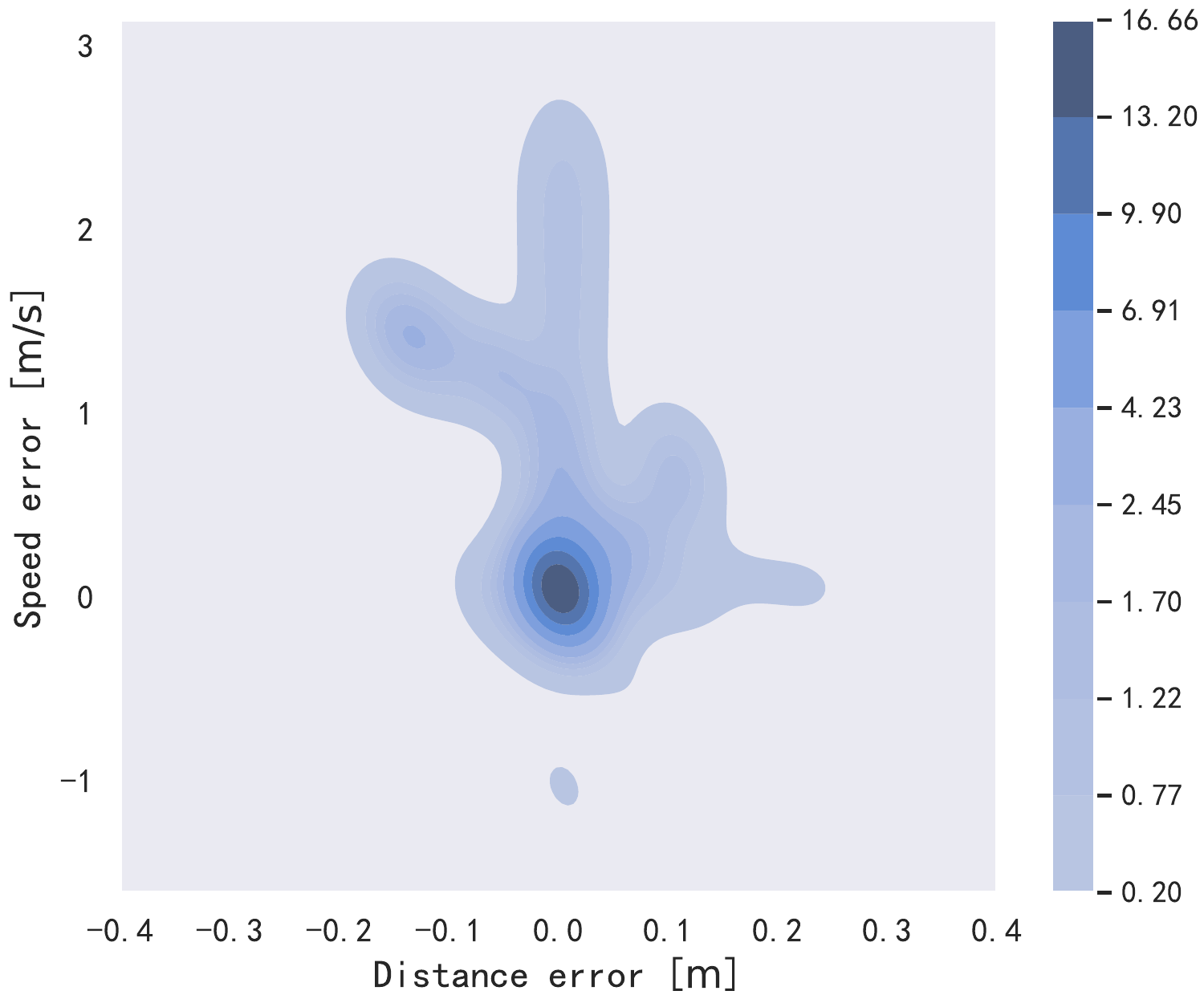}}
\subfloat[Control commands distribution]{\label{control_action_dist}\includegraphics[width=0.5\linewidth]{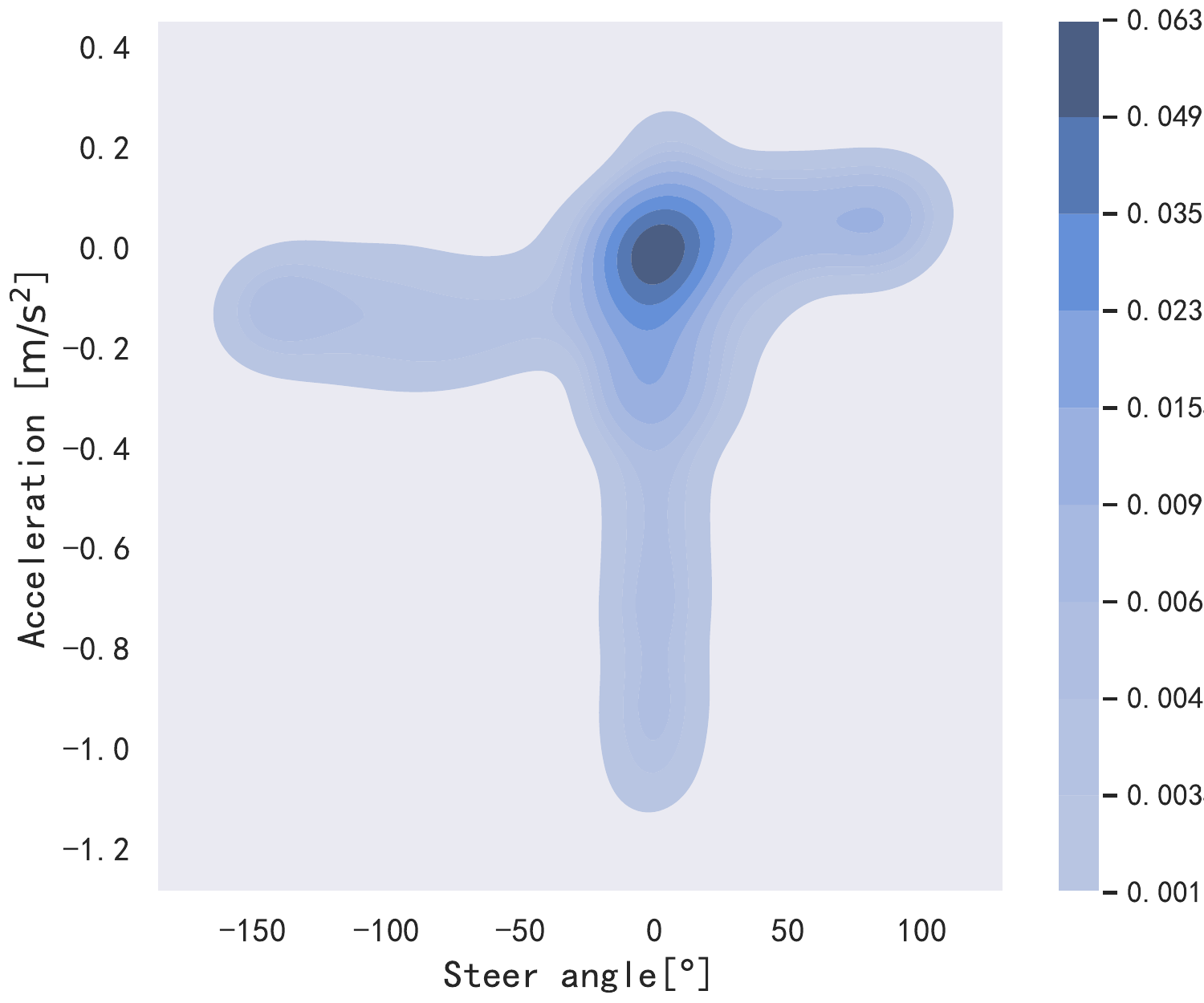}}
\caption{Tracking performance of diversified paths.}
\label{f:tracking performance}
\end{figure}

To verify the rationality of the paths generated by the general static path planner, we firstly test the path tracking function with the trained policy where we remove all traffic flows at the intersections plotted in Fig.~\ref{fig.exp_route_planning}. 
We conduct 200 runs totally based on the six distinct intersections and record the tracking error and vehicle state of each step. We randomly sample the route and velocity mode for each run and always initialize the automated vehicle outside the intersection, then the trained policy will drive it to track the given path. The distribution of the tracking error and vehicle state as shown in Fig \ref{f:tracking performance}. Note that the distance error is defined by $\Delta x$ and the longitudinal error $\Delta y$, i.e., $\sqrt{\Delta x ^2 + \Delta y ^2}$.
We can see that the tracking error is close to zero in most cases with the maximum as 0.2m. And the maximum speed error is less than 3m/s, and also mainly concentrates around 0.
Fig.~\ref{control_action_dist} demonstrate that steer angle and acceleration are close to zero in most cases, of which the steering angle keeps less than $150^o$ and the absolute value of acceleration keeps less than $1.2\rm m/s^2$. In addition, there is no situation where the two variables are large simultaneously, which means the vehicle stability can be guaranteed. To sum up, the general static path planner can generate reasonable candidate paths for diversified intersections, which are easily trackable for the automated vehicle with an acceptable energy consumption.

\begin{figure*}[!htbp]
\centering
\captionsetup[subfigure]{justification=centering}
\subfloat[t=1.5s]{\label{f:demo_step15}\includegraphics[width=0.24\linewidth]{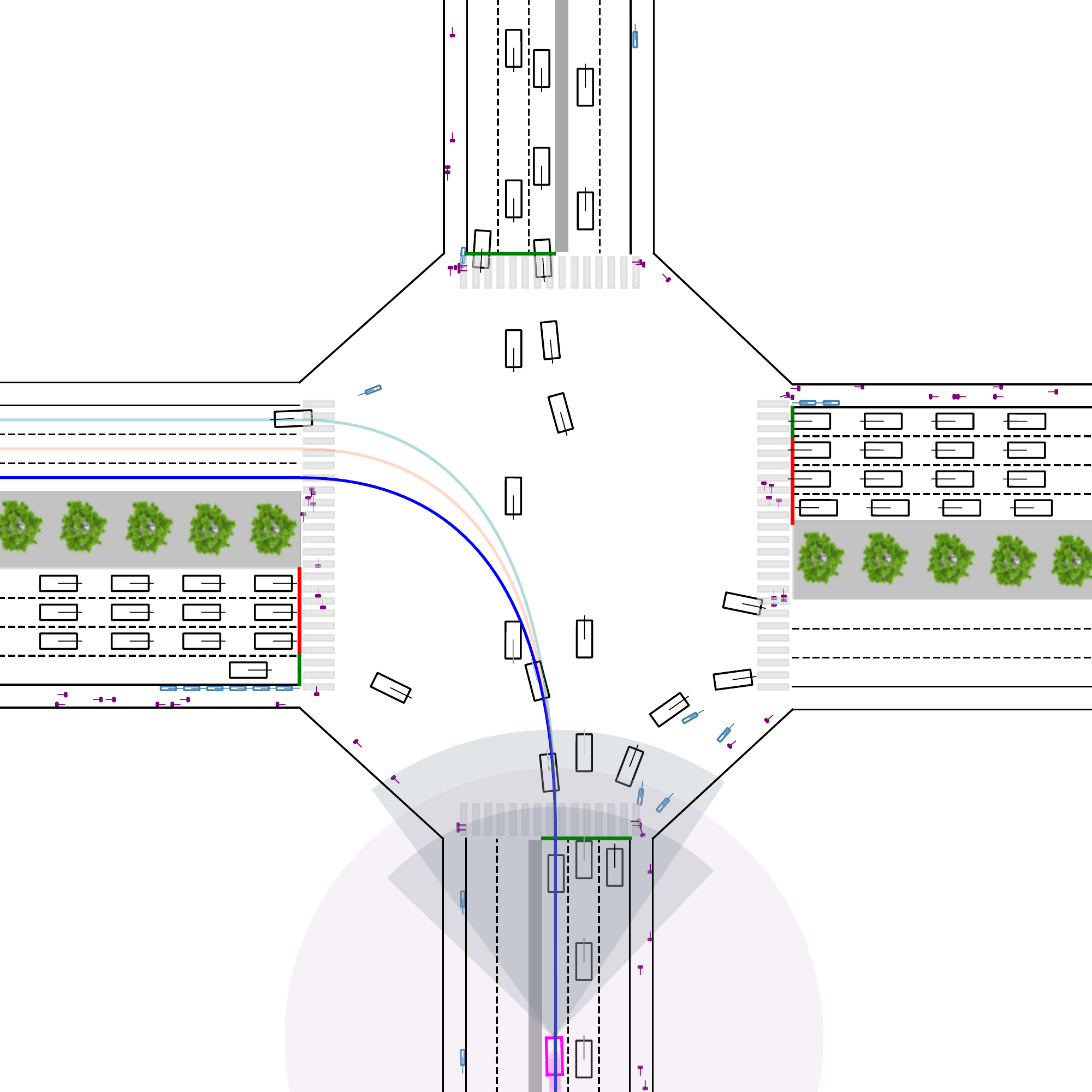}} \
\subfloat[t=3.6s]{\label{f:demo_step36}\includegraphics[width=0.24\linewidth]{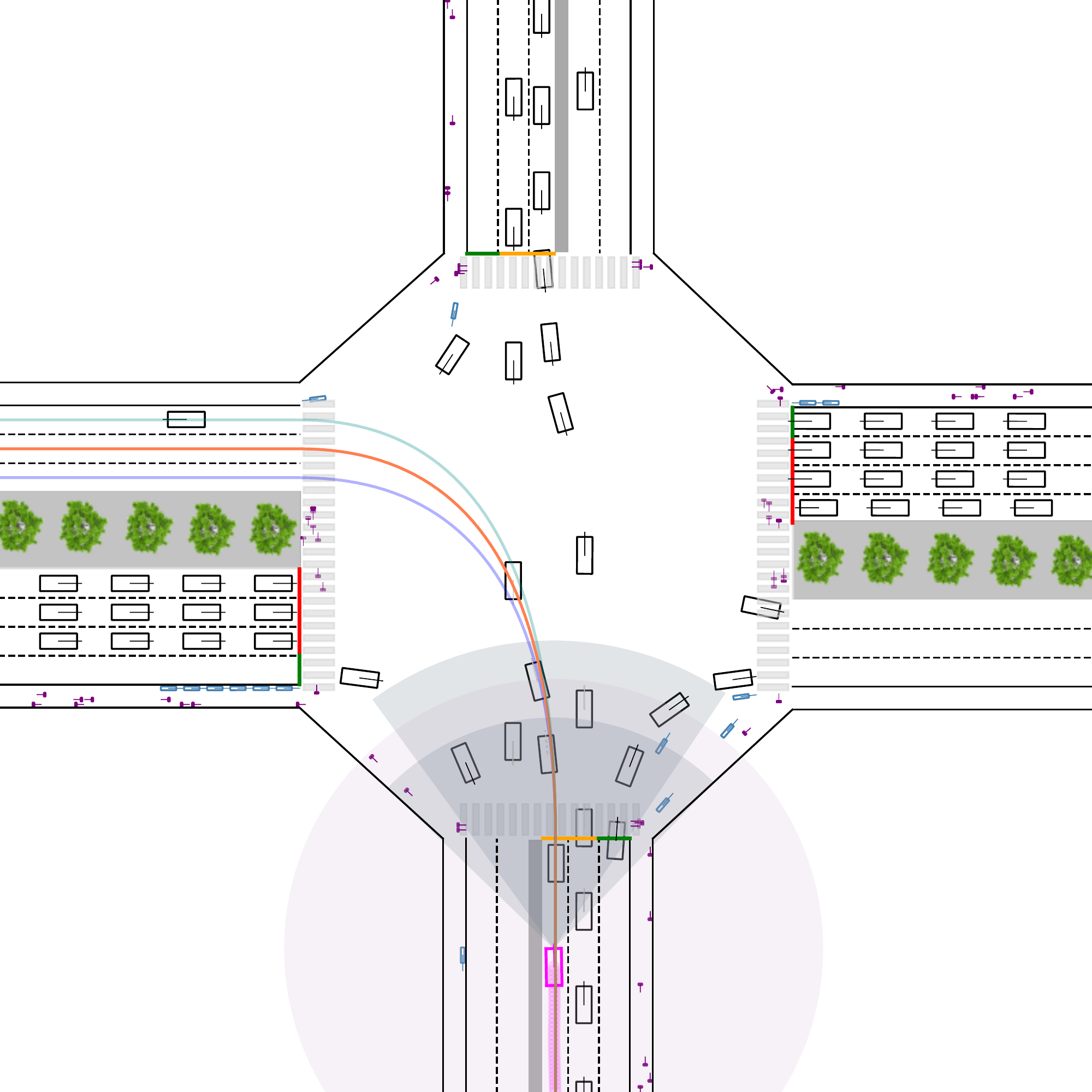}} \
\subfloat[t=7.0s]{\label{f:demo_step70}\includegraphics[width=0.24\linewidth]{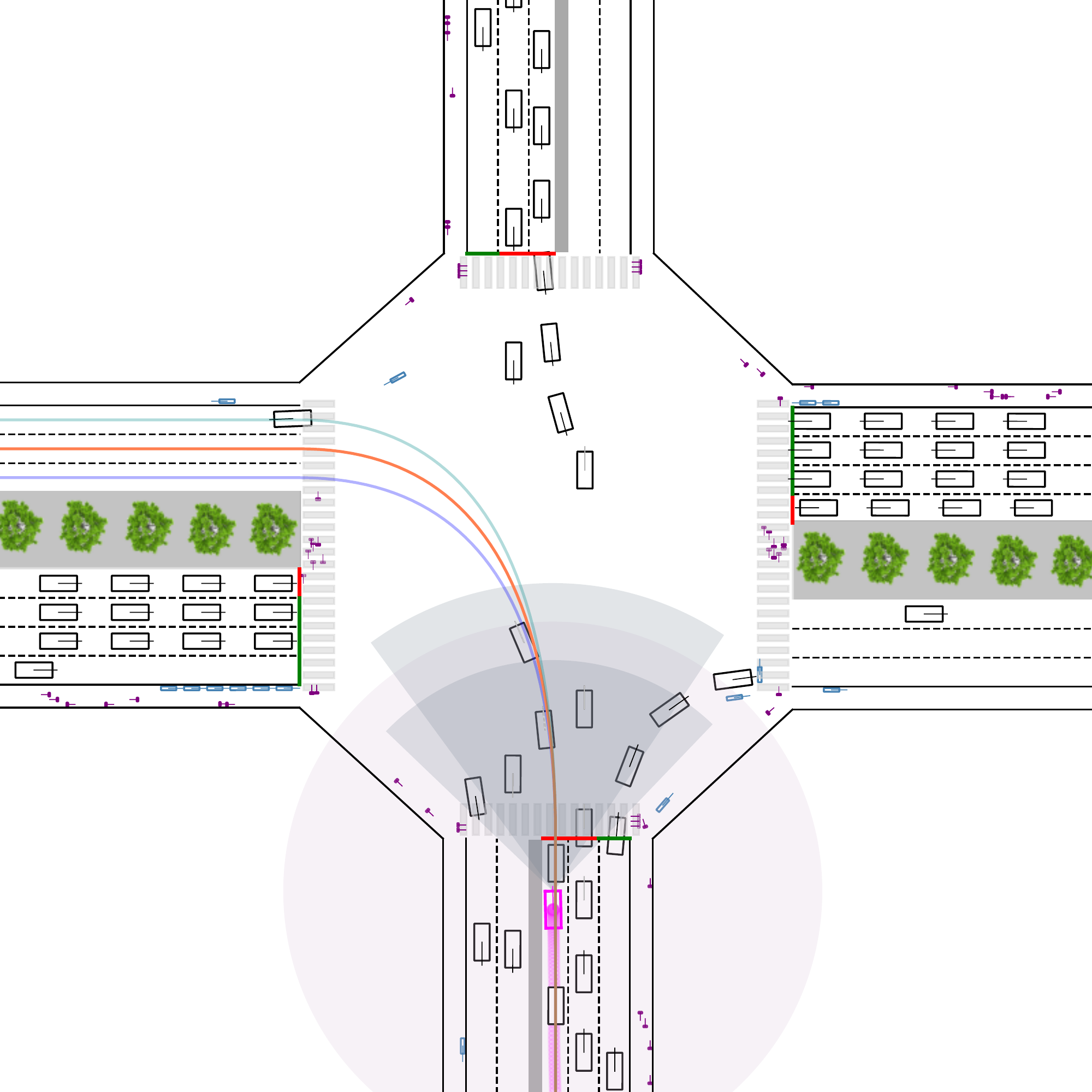}} \
\subfloat[t=72.5s]{\label{f:demo_step72}\includegraphics[width=0.24\linewidth]{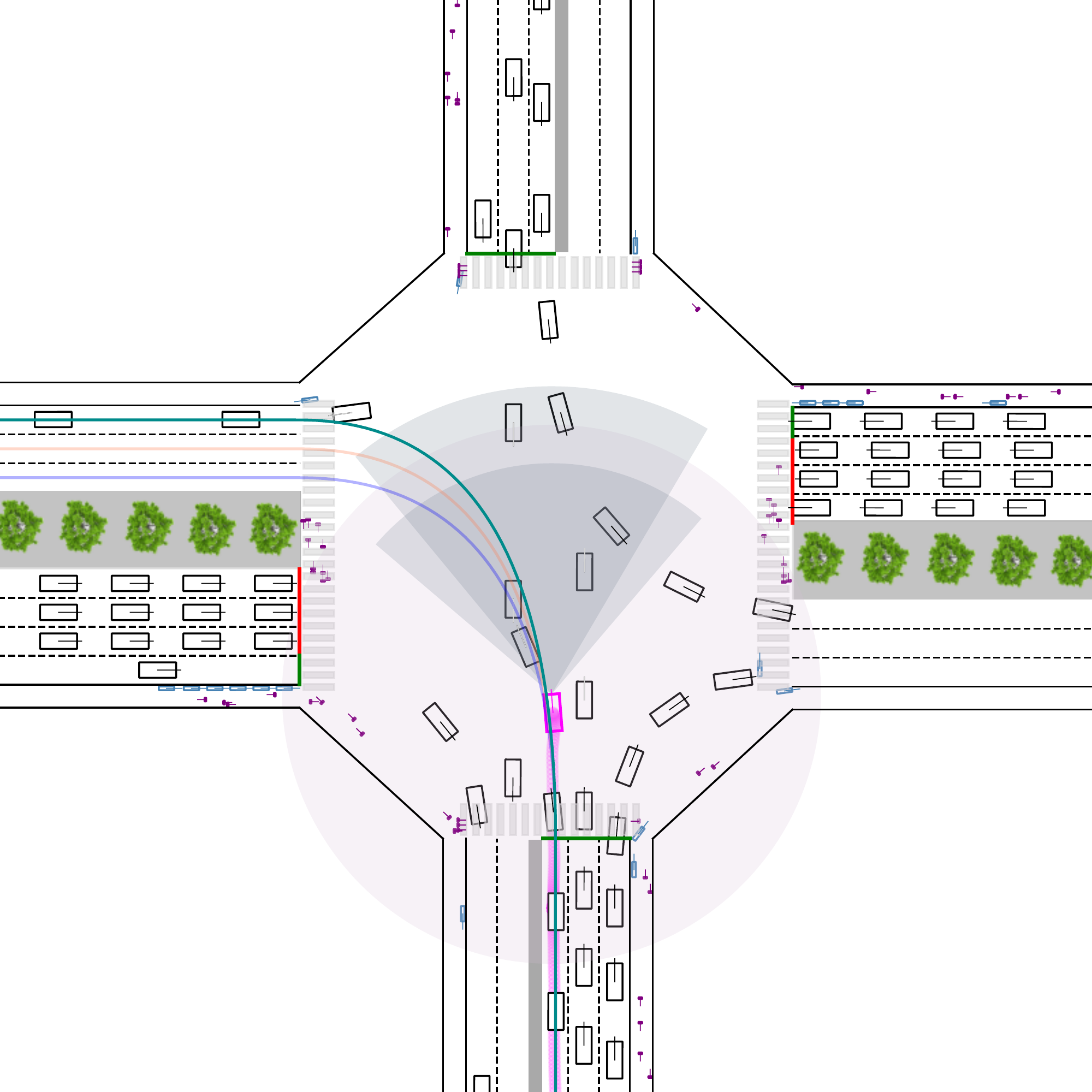}}
\\
\subfloat[t=78.5s]{\label{f:demo_step78}\includegraphics[width=0.24\linewidth]{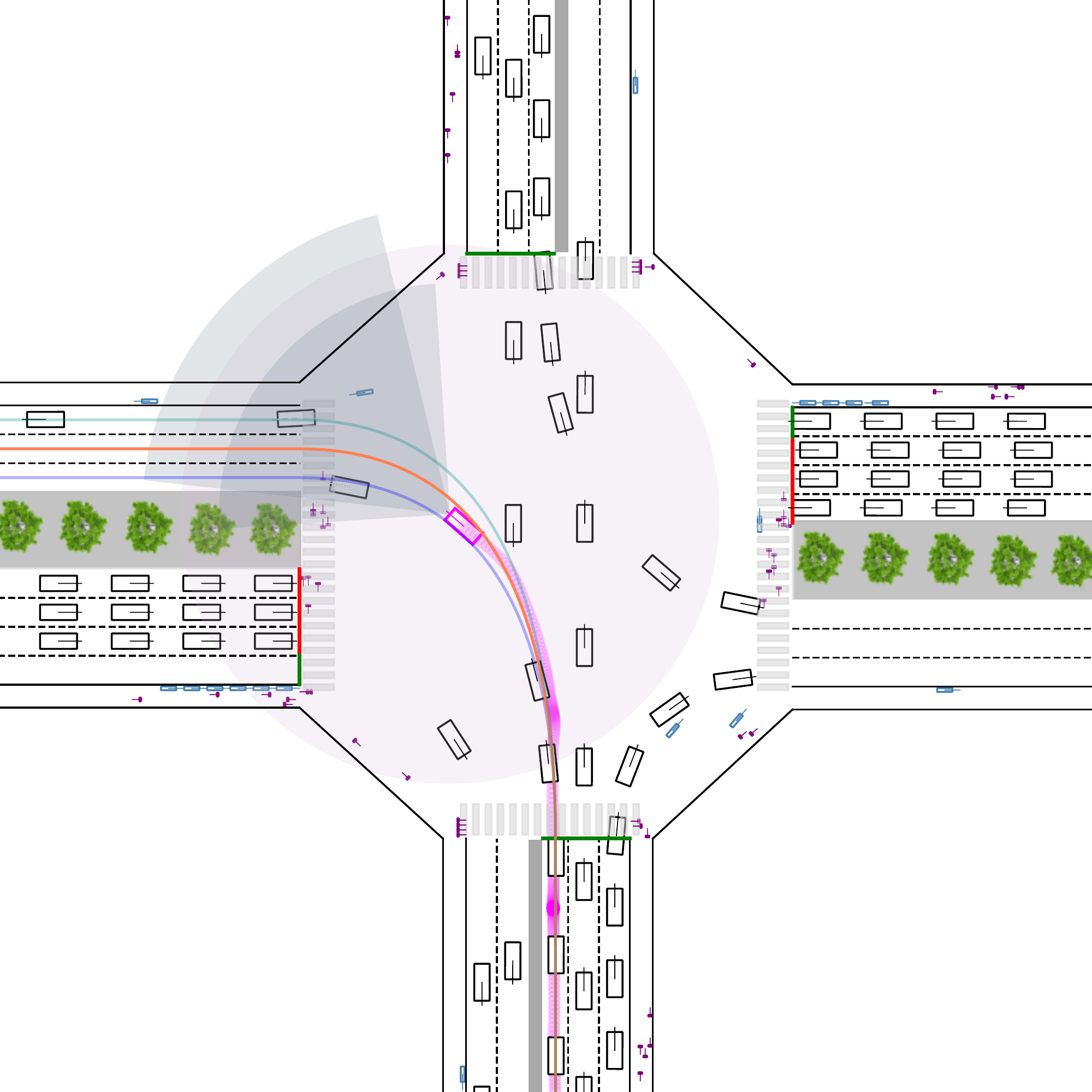}} \
\subfloat[t=82.0s]{\label{f:demo_step82}\includegraphics[width=0.24\linewidth]{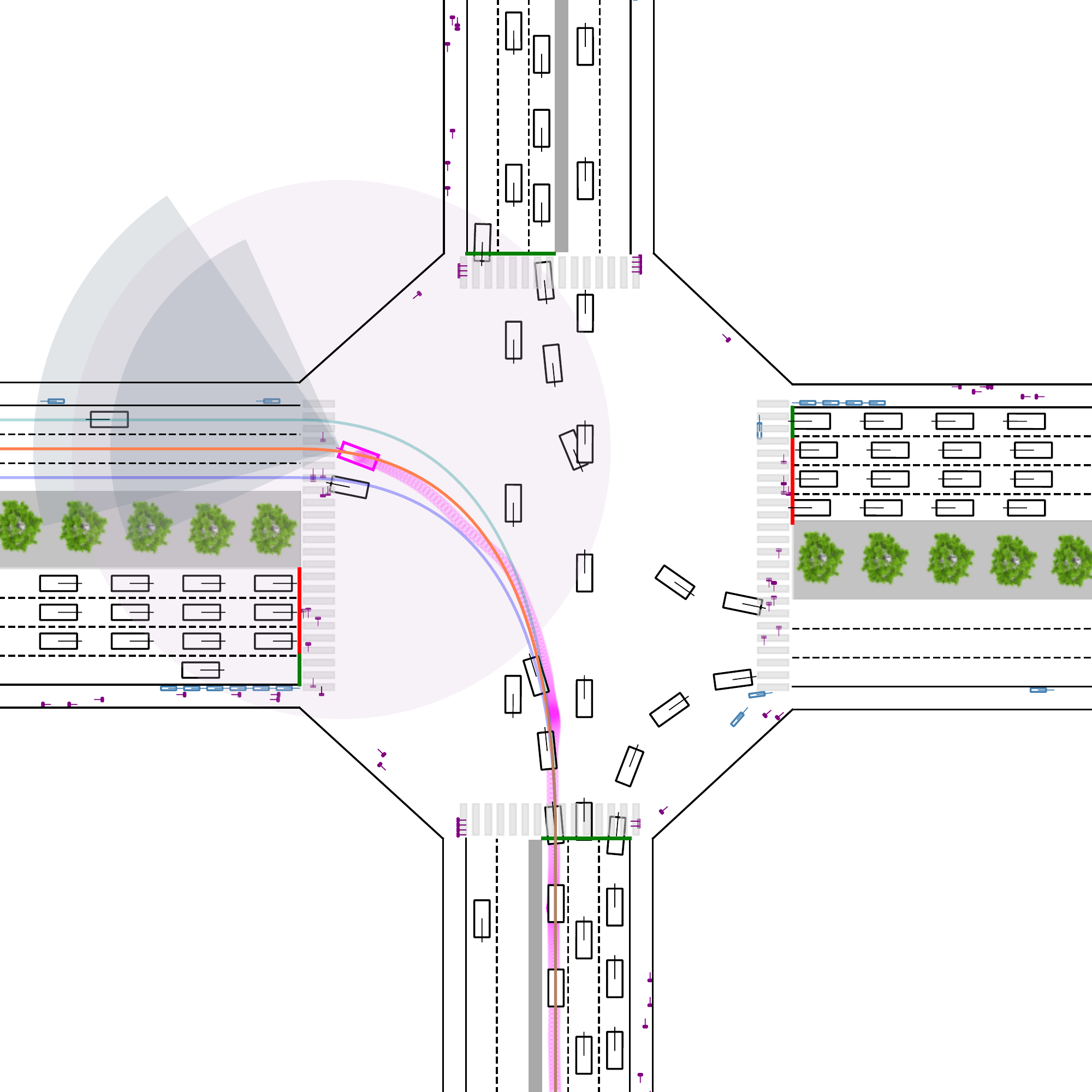}} \
\subfloat[t=85.9s]{\label{f:demo_step85}\includegraphics[width=0.24\linewidth]{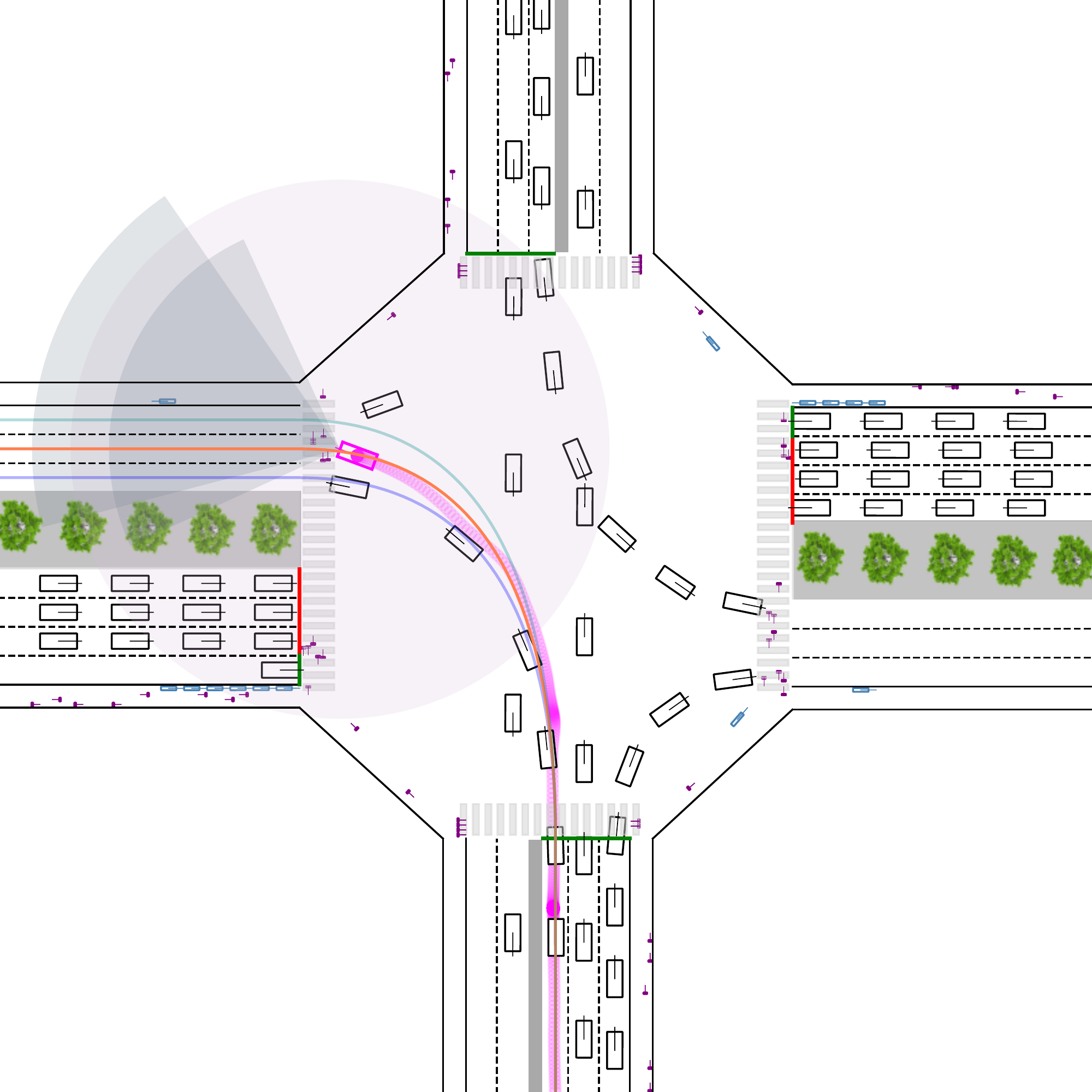}} \
\subfloat[t=94.0s]{\label{f:demo_step94}\includegraphics[width=0.24\linewidth]{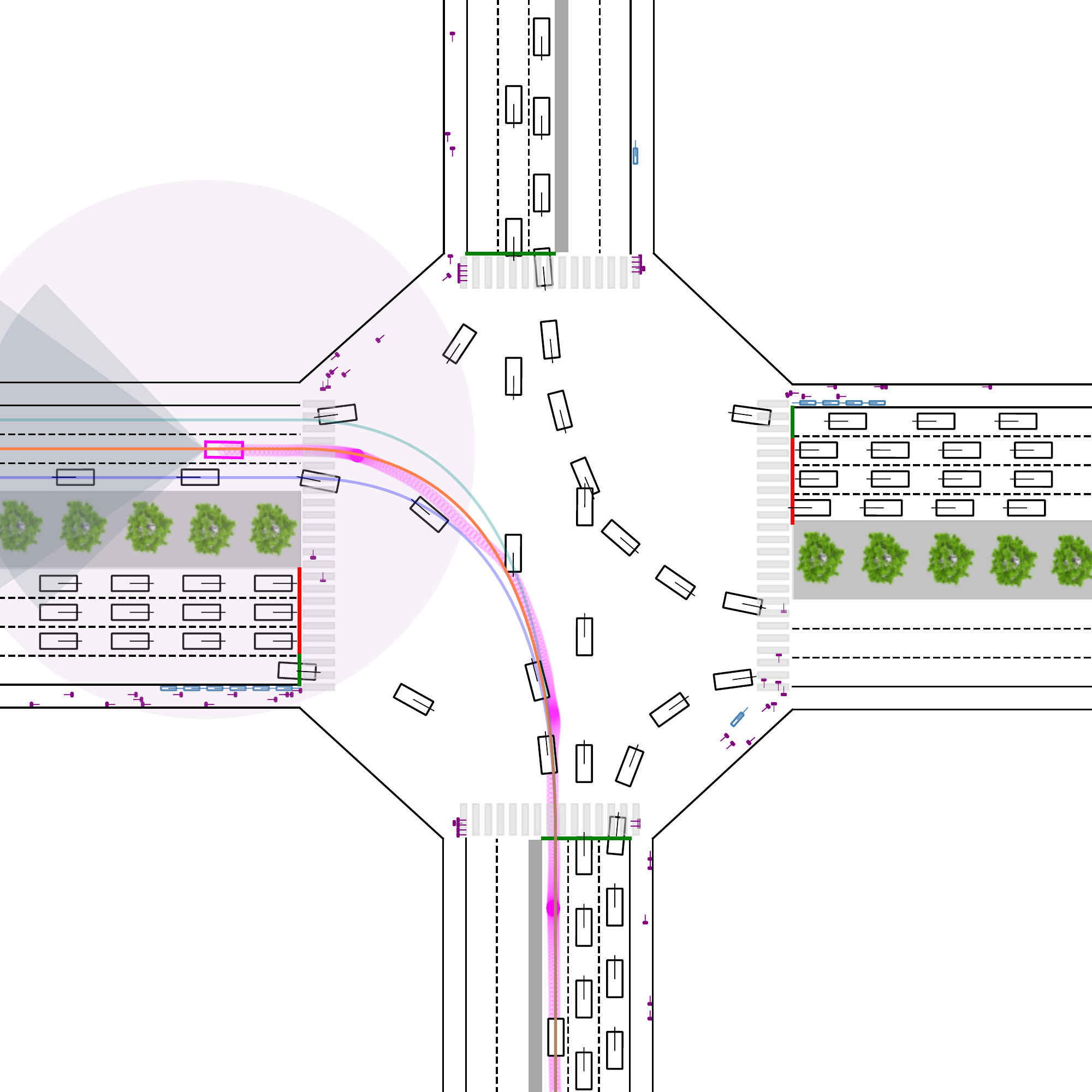}} \
\caption{Trajectory visualization for the left-turn task. The red box represents the ego vehicle controlled by the trained policy and the highlighted curve is the optimal path chosen by value function. The black, blue and purple rectangles indicate the surrounding vehicles, cyclists and pedestrians respectively. The shaded sectors are the perception range of radar, camera and lidar, and the red pots trailed behind the automated vehicle shows its history trajectory until this moment.}
\label{fig.simu_traj}
\end{figure*}

\begin{figure}[!htb]
\captionsetup{justification =centering,
              singlelinecheck = false,labelsep=period, font=small}
\subfloat[Speed mode]{\label{f:demo_route}\includegraphics[width=1.0\linewidth]{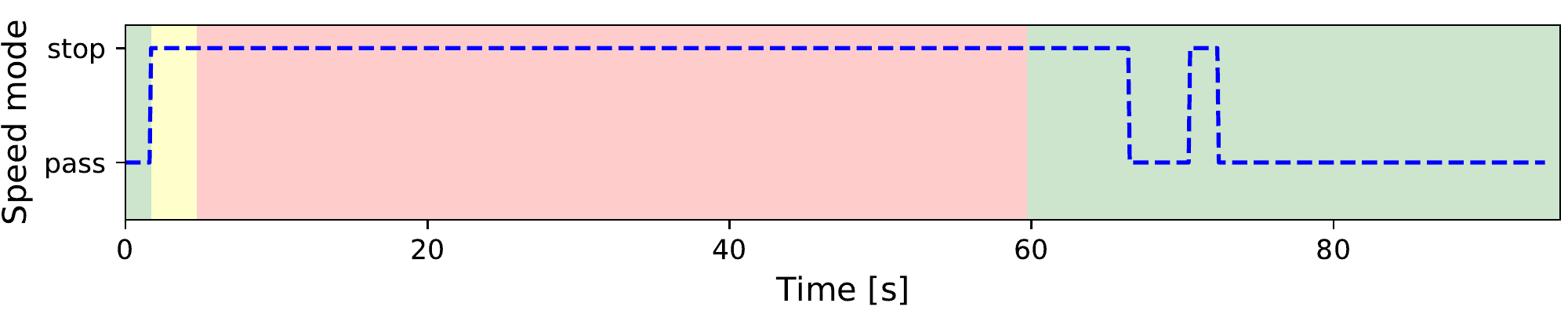}} \\
\subfloat[Speed]{\label{f:demo_speed}\includegraphics[width=1.0\linewidth]{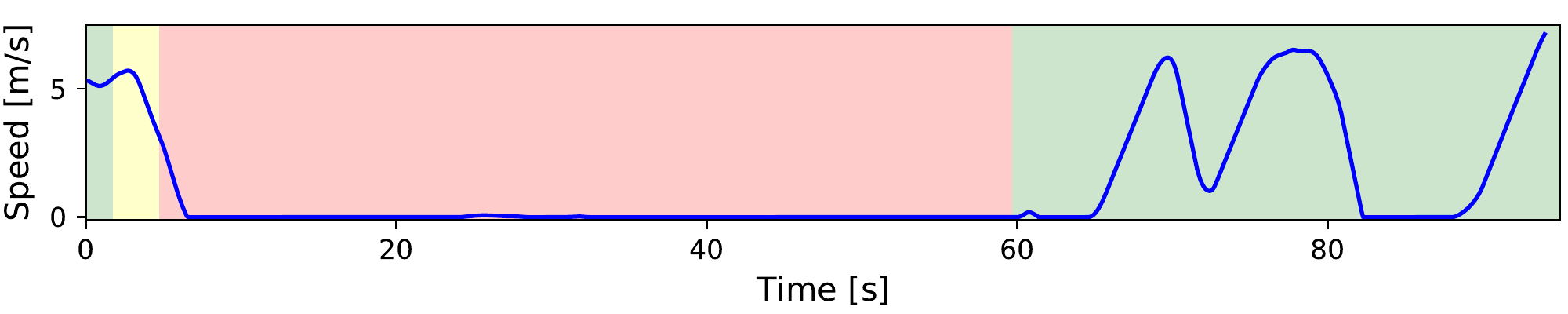}}  \\
\subfloat[Steer angle]{\label{f:demo_steer}\includegraphics[width=1.0\linewidth]{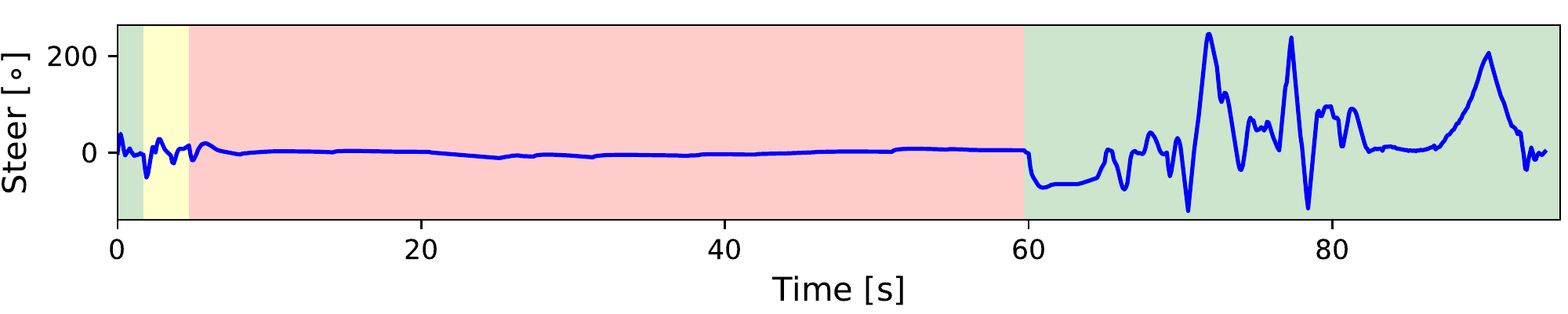}}  \\
\subfloat[Acceleration]{\label{f:demo_acc}\includegraphics[width=1.0\linewidth]{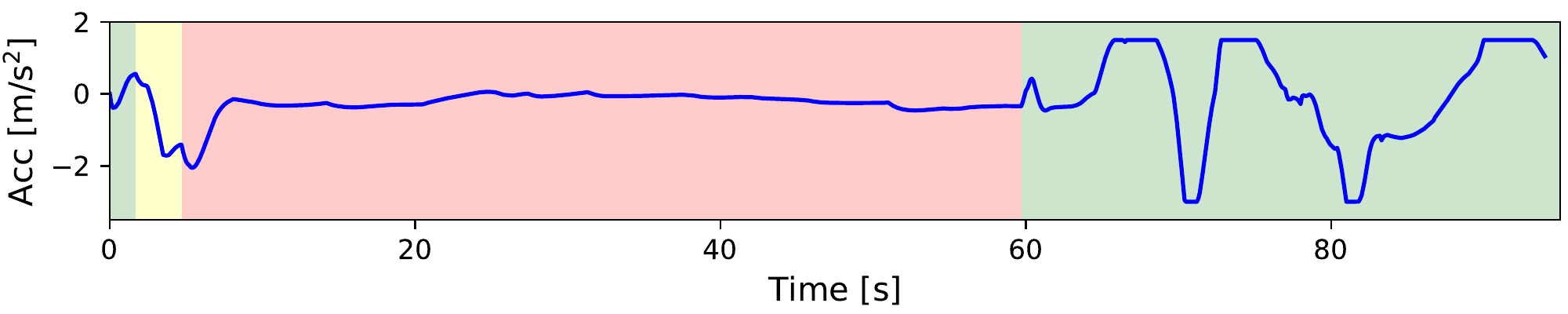}}
\caption{State curves of the left-turn task. The shadow background of these figures indicate the signal light phase, i.e., greed, red, yellow respectively.}
\label{f:avg_speed}
\end{figure}

Next, we visualize a typical driving process of the unprotected left-turn task and the curves of some important vehicle states in Fig.~\ref{fig.simu_traj} and Fig.~\ref{f:avg_speed}, which is considered as one of the most difficult tasks at signalized intersections. The automated vehicle starts from the left-turn lane with the green light shown in Fig.~\ref{f:demo_step15}, wherein the pass velocity in Fig.~\ref{f:demo_route} and the bottommost route  will be chosen to drive towards the intersection as fast as possible. Then the signal light goes into the yellow phase and immediately the stop velocity mode works to decelerate the ego as there is a long distance from the stop line and also a front vehicle attempts to stop as shown in  Fig.~\ref{f:demo_step36}. After that, the ego vehicle continues deceleration and stops to wait the red light in Fig.~\ref{f:demo_step70} and Fig.~\ref{f:demo_speed}.
Next, although the green light appears at 60s in Fig.~\ref{f:demo_route}, the stop velocity still makes the difference because the front vehicle yields to other vehicles stuck inside the intersection. Later, the ego vehicle will
choose the pass mode and accelerate to a high speed as shown in Fig.~\ref{f:demo_acc} and it shall detour the sluggish front vehicle by choosing the topmost path as shown in Fig.~\ref{f:demo_step72}, which needs to steer right for a while as shown in Fig.~\ref{f:demo_steer}. After that it will quickly switch to the middle path in Fig.~\ref{f:demo_step78} to rush before the opposite straight-going vehicle. Soon, our automated vehicle drives close to the sidewalk in Fig.~\ref{f:demo_step85} through which some pedestrians are walking. Thus, it will decelerate priorly and yield to pedestrians as shown in Fig.~\ref{f:demo_step82} and Fig.~\ref{f:demo_speed}.
Finally, the automated vehicle starts to accelerate again at 86s and pass this intersection successfully as demonstrated in Fig.~\ref{f:demo_step94} and Fig.~\ref{f:demo_acc}, during which it prefers the middle path to track because the sparse participants in this lane will reduce the potential collisions.

\begin{figure}[!htbp]
\centering
\captionsetup[subfigure]{justification=centering}
\subfloat[t=0.0s]{\label{f:simu2_step00}\includegraphics[width=0.45\linewidth]{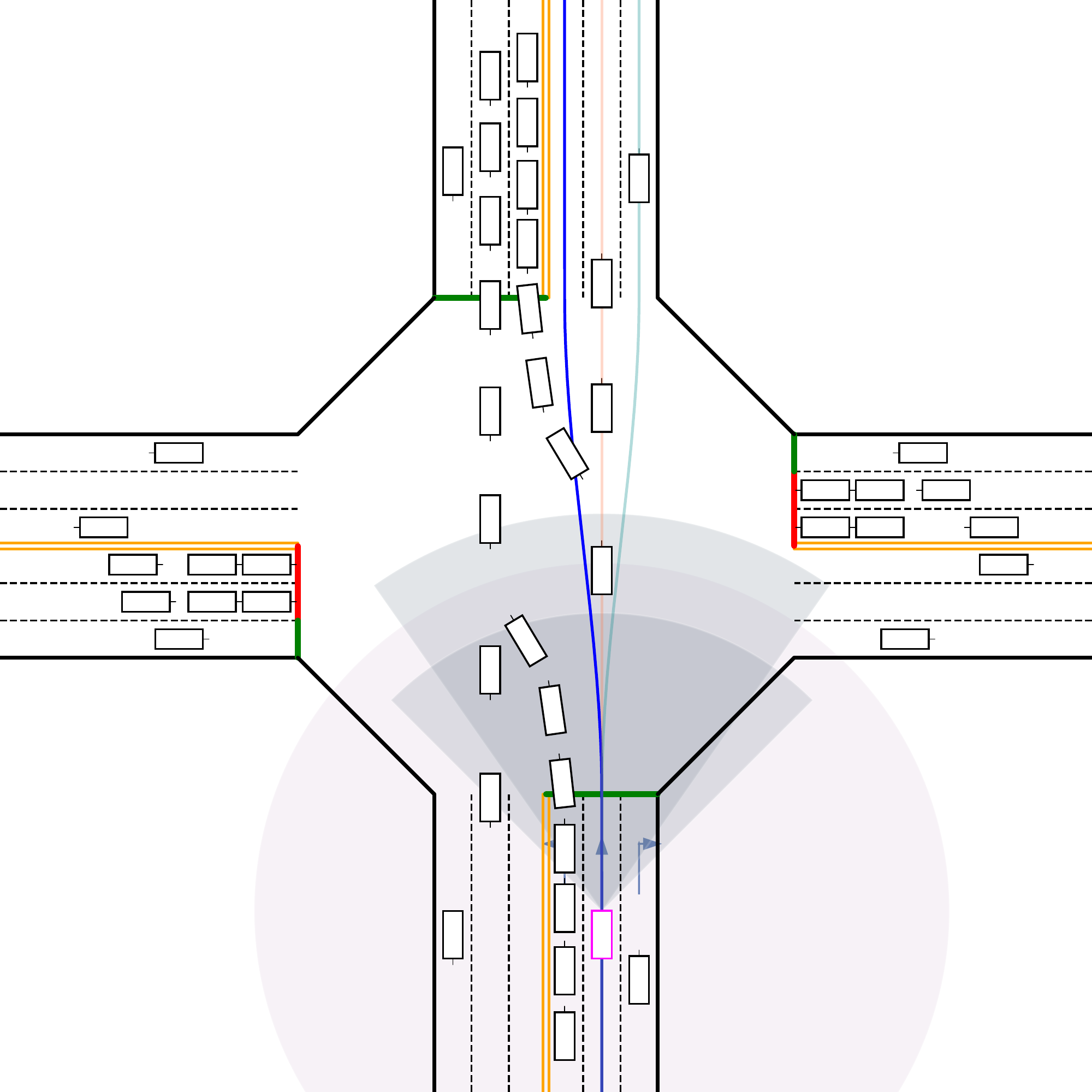}} \
\subfloat[t=4.0s]{\label{f:simu2_step40}\includegraphics[width=0.45\linewidth]{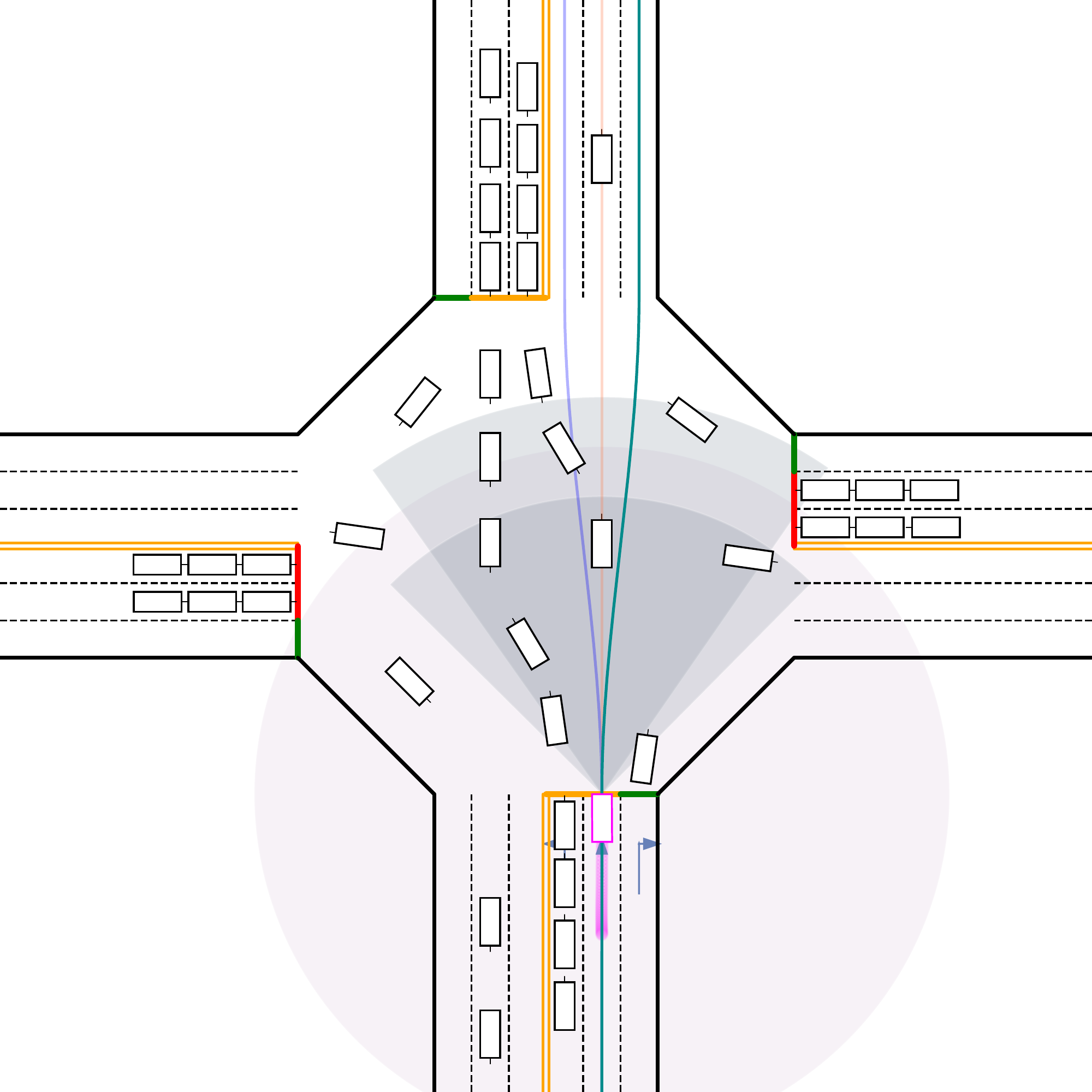}} \
\\
\subfloat[t=10.0s]{\label{f:simu2_step100}\includegraphics[width=0.45\linewidth]{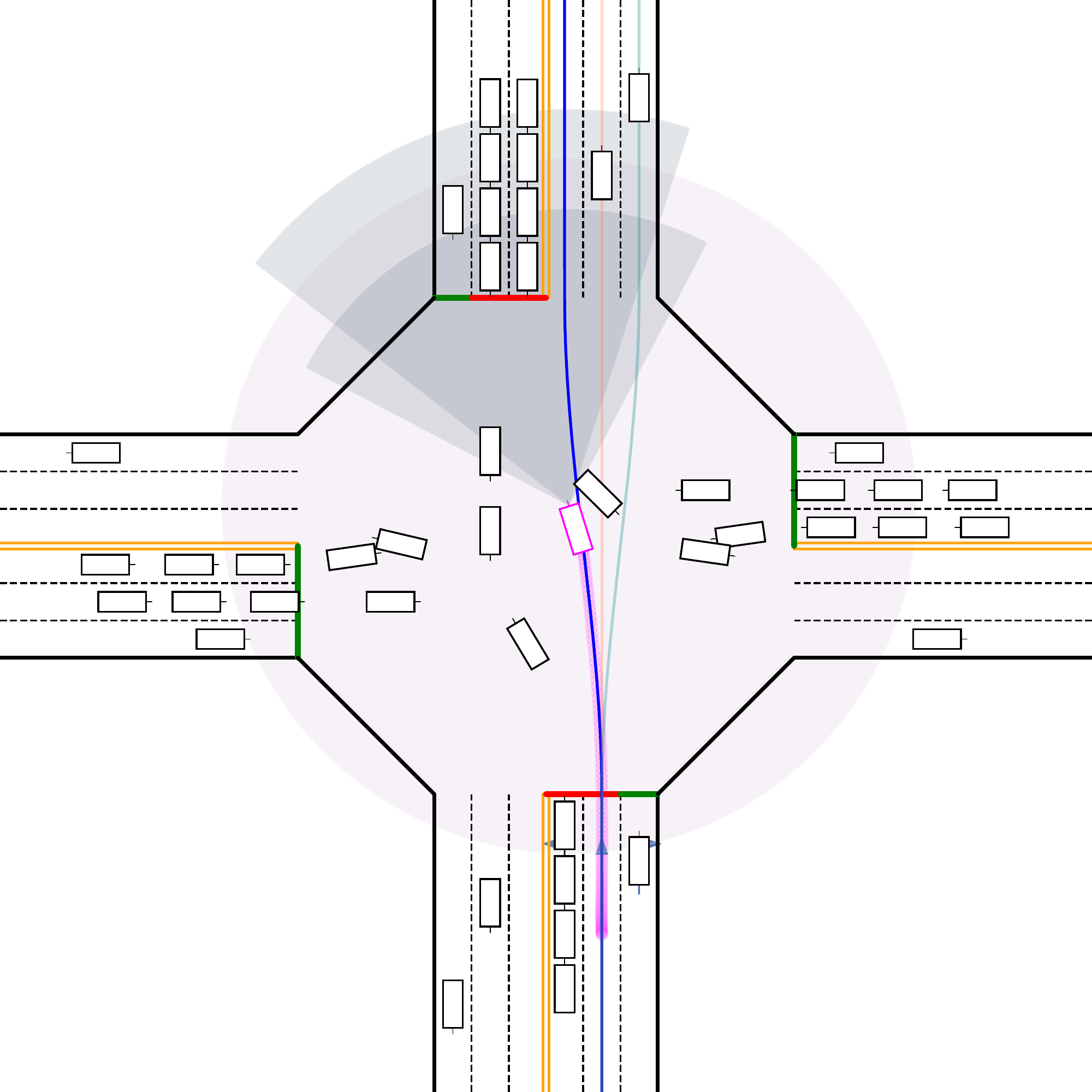}} \
\subfloat[t=17.2s]{\label{f:simu2_step172}\includegraphics[width=0.45\linewidth]{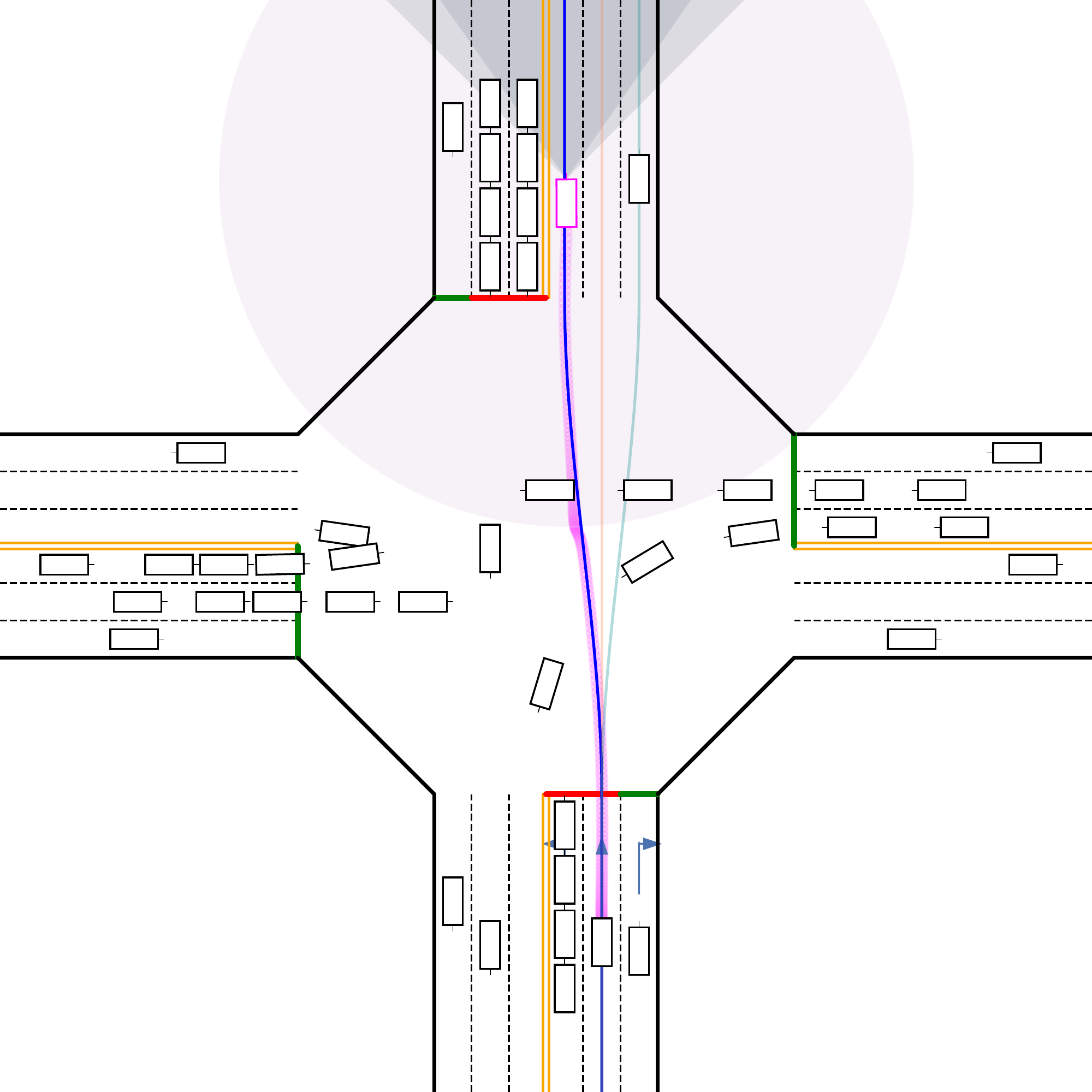}} \
\caption{Trajectory visualization for the straight-going task.}
\label{fig.simu_traj_straight}
\end{figure}

\begin{figure}[!htbp]
\centering
\captionsetup[subfigure]{justification=centering}
\subfloat[t=0.0s]{\label{f:simu3_step00}\includegraphics[width=0.45\linewidth]{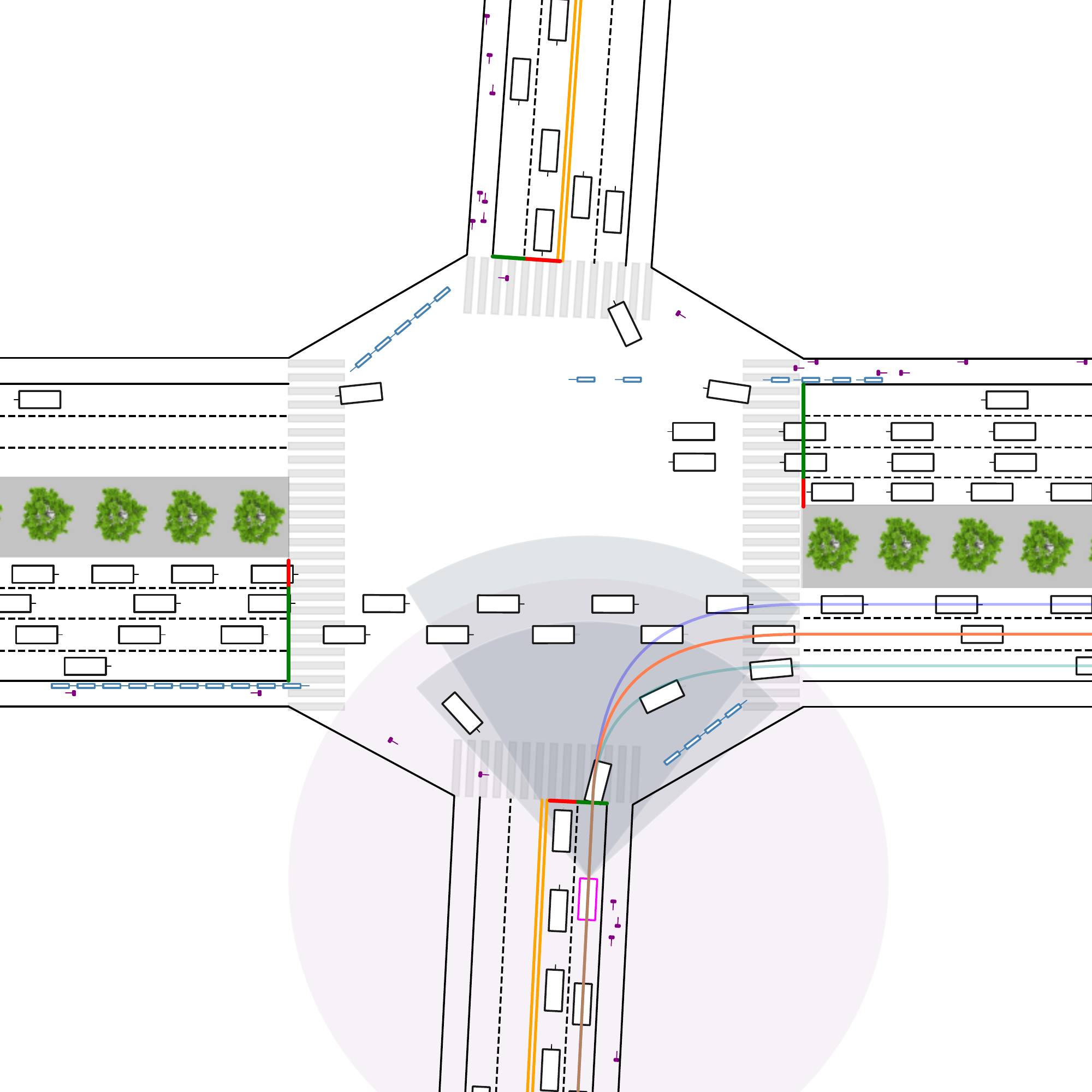}} \
\subfloat[t=4.0s]{\label{f:simu3_step35}\includegraphics[width=0.45\linewidth]{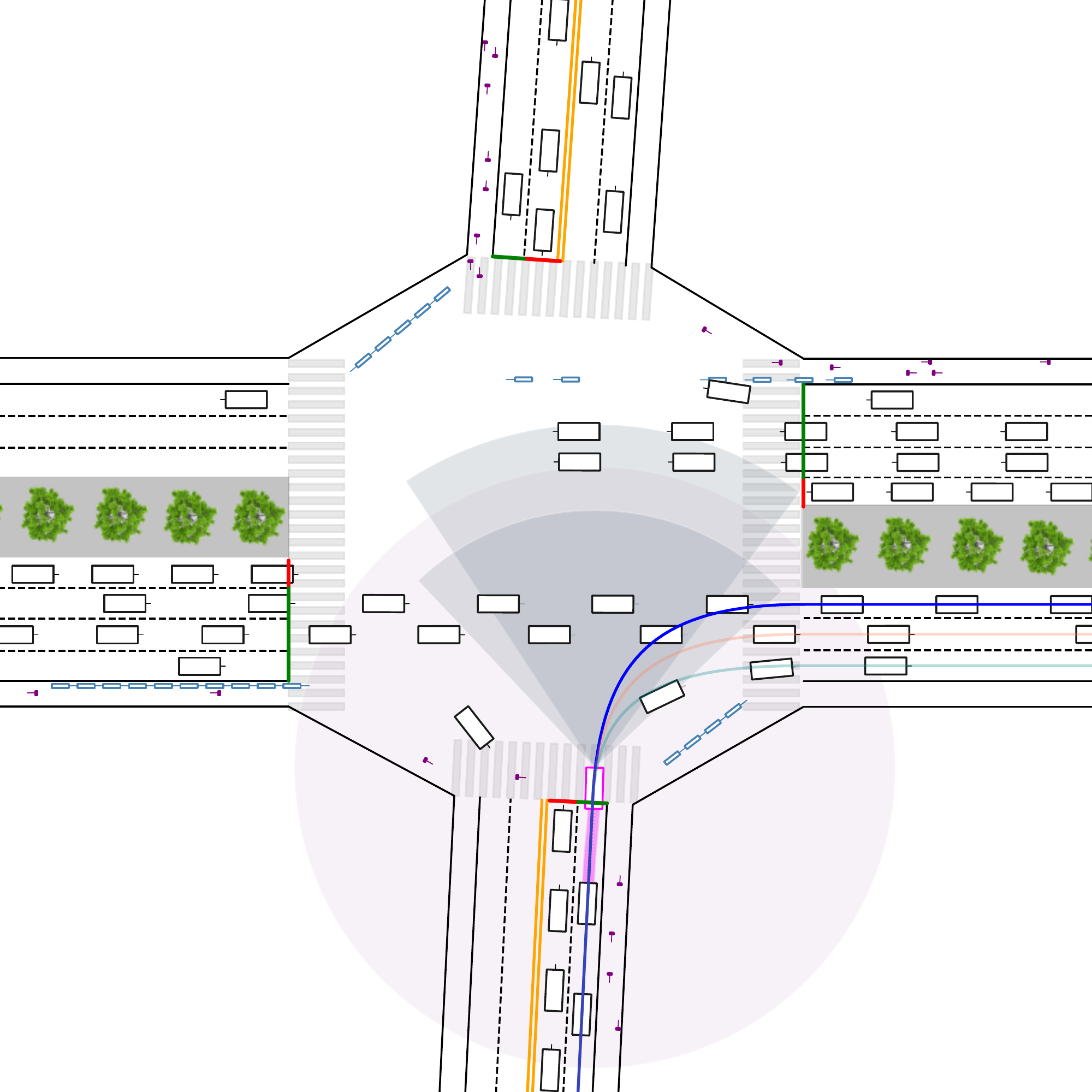}} \
\\
\subfloat[t=10.0s]{\label{f:simu3_step110}\includegraphics[width=0.45\linewidth]{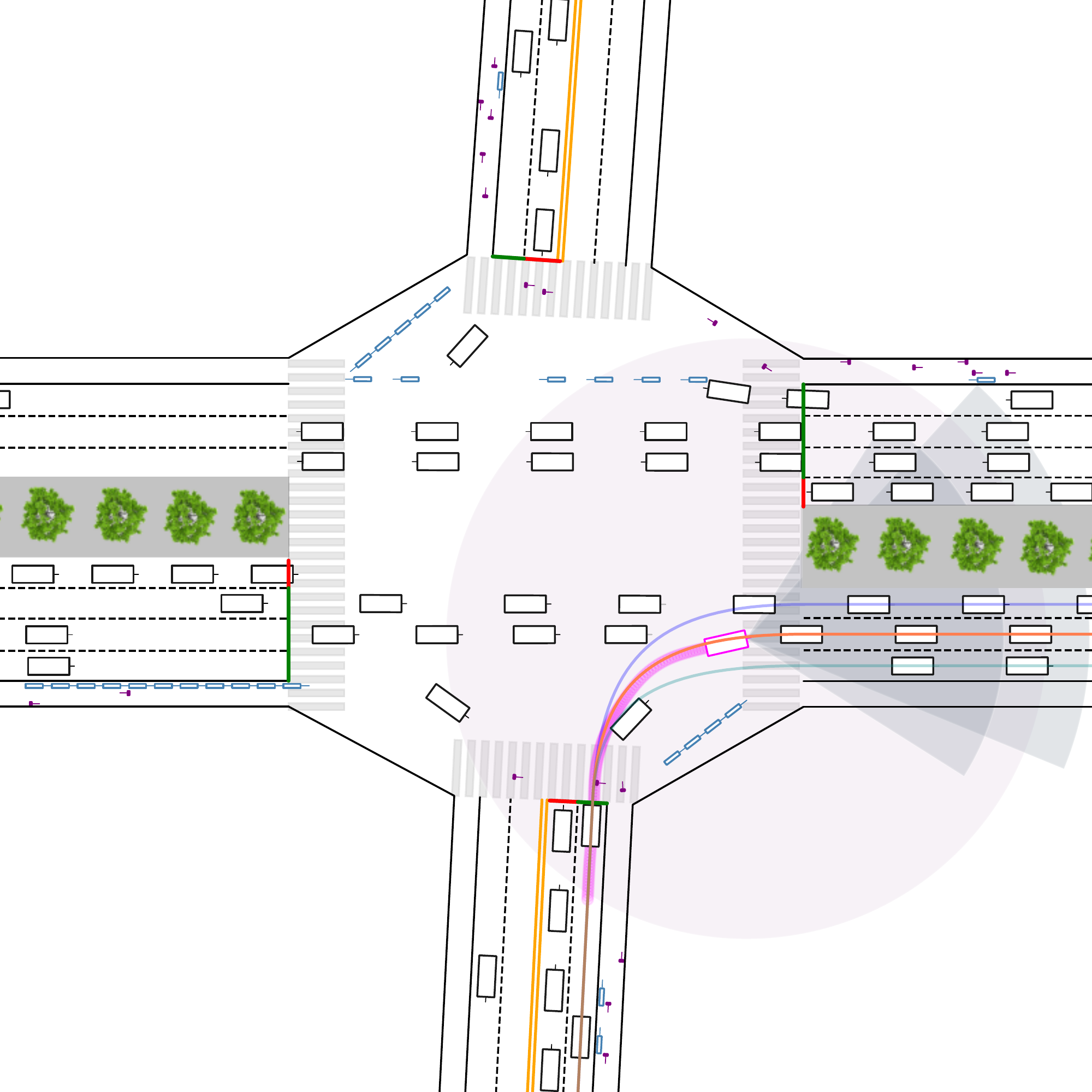}} \
\subfloat[t=17.2s]{\label{f:simu3_step150}\includegraphics[width=0.45\linewidth]{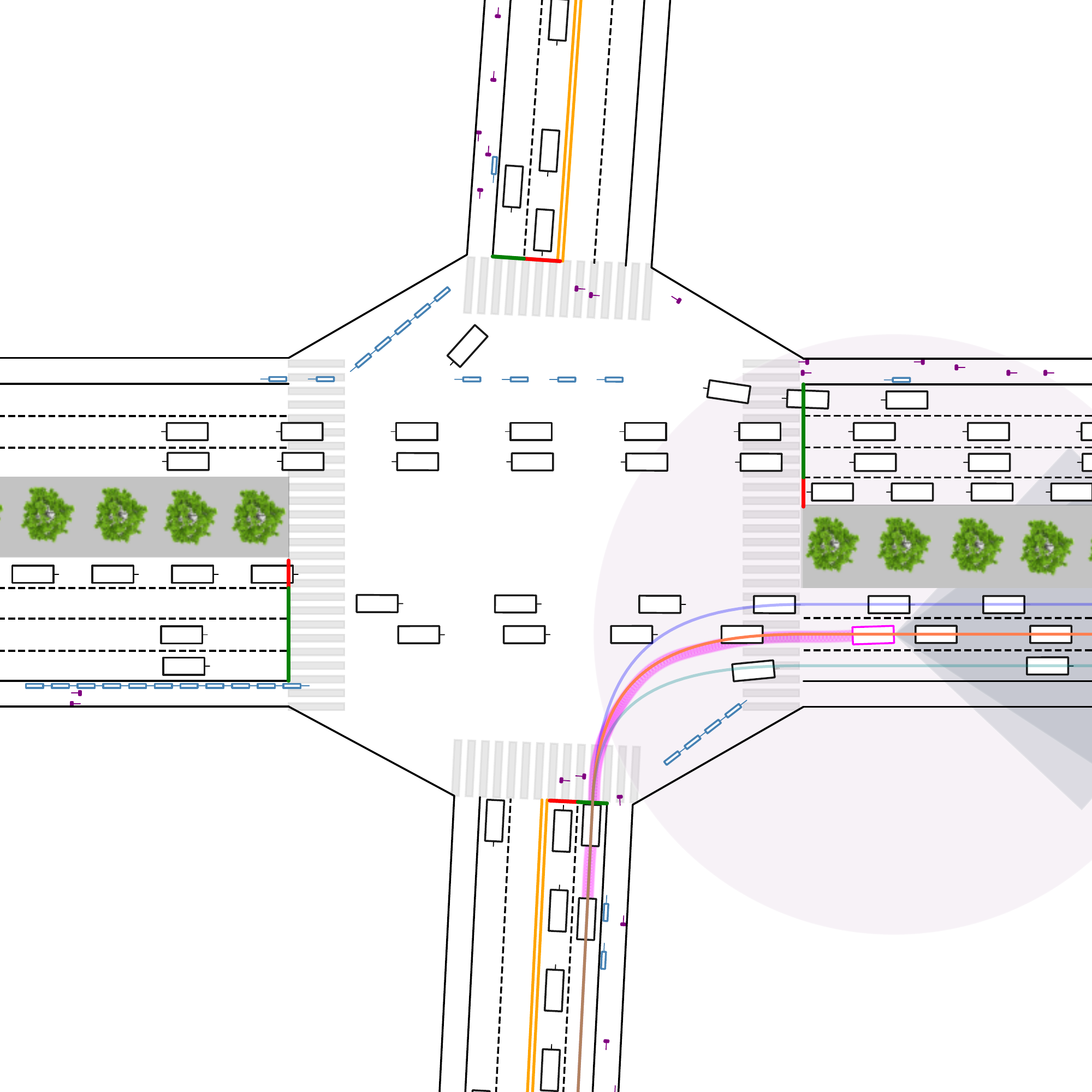}}
\caption{Trajectory visualization for the right-turn task.}
\label{fig.simu_traj_right}
\end{figure}

Besides, we also visualize the straight-going and right-turn tasks at two other intersections.
In Fig.~\ref{fig.simu_traj_straight}, the ego vehicle starts initially close to the stop line at green light in Fig.~\ref{f:simu2_step00} and thus it will disregard the yellow light by choosing the pass velocity in Fig.~\ref{f:simu2_step40}. After entering into the intersection, it will steer left to avoid the opposite turning-left vehicle by selecting the leftmost route in Fig.~\ref{f:simu2_step100} and finally finish the safe navigation at this intersection shown as Fig.~\ref{f:simu2_step172}.
In Fig.~\ref{fig.simu_traj_right}, the ego vehicle is initialized in the right-turning lane in Fig.~\ref{f:simu3_step00} with the constant green lights. Then it will acceleration to pass before the upcoming pedestrians in Fig.~\ref{f:simu3_step35} and attempt to avoid the front vehicle by choosing the middle path in Fig.~\ref{f:simu3_step110}. Eventually, it will insert the straight-going traffic flow to pass successfully in Fig.~\ref{f:simu3_step150}.

\subsection{Generalization ability}
\begin{figure}[!htb]
\captionsetup{singlelinecheck = false,labelsep=period, font=small}
\centering
\captionsetup[subfigure]{justification=centering}
\subfloat[The 1st test case]{\label{f:1st_test_case}\includegraphics[width=0.45\linewidth]{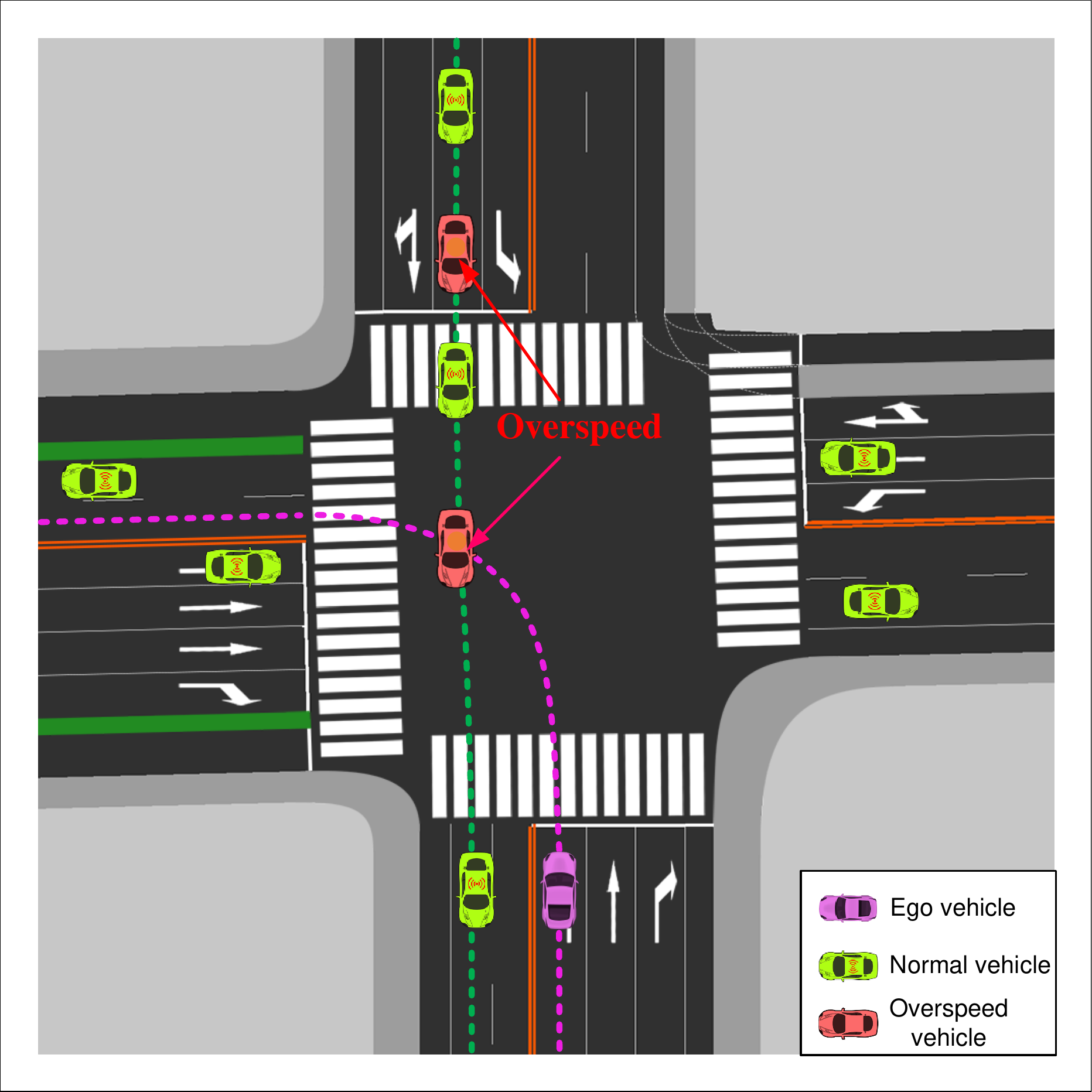}}
\subfloat[The 2nd test case]{\label{f:2nd_test_case}\includegraphics[width=0.45\linewidth]{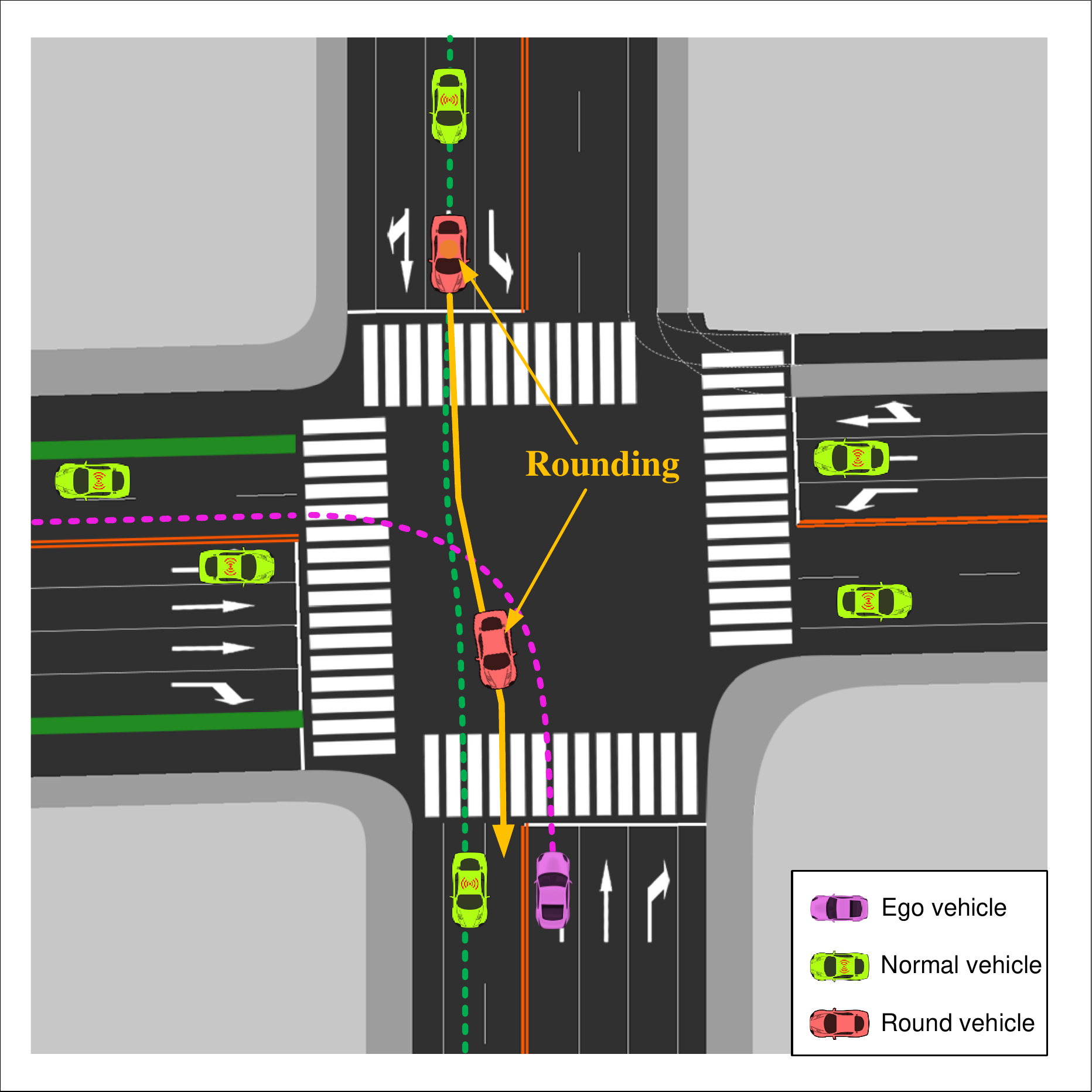}}
\caption{Test case design to evaluate generalization ability. (a) The red surrounding vehicle at the straight-going lane will overspeed to rush through. (b) The red surrounding vehicle at the straight-going lane will round into the inside lane.}
\label{f:test_case}
\end{figure}

Compared with the performance during training process, we also concern more about that on situations distinct from the training environment, for example, these may exit some abnormal behaviors of the surrounding participants which rarely encounters during training.
Here, we design two typical conditions wherein the part of surrounding vehicles are endowed with some abnormal driving behaviors to test the generalization ability.
Our tests will focus on the unprotected left-turn task with green light set as green.
In the first case shown in Fig.~\ref{f:1st_test_case}, 15\% vehicles in the opposite straight lane will overspeed to rush through the intersections as quickly as possible.
That is, the speed limits of the surrounding vehicles are set as 25km/h for the roads inside of the training intersection set and 30km/h for the outsides. However, during the testing, part of surrounding vehicles will exceed the speed limits by 10\%, 20\% and 50\% respectively.
As for the second case shown in Fig.~\ref{f:2nd_test_case}, a proportion of vehicles of the traffic flow will violate the road connection, and attempt to round and enter into the inside lane wherein the middle lane is only allowed to ride during training.
The proportion will be set to 10\%, 20\% and 50\% respectively.
Two crucial indicators including passing rate and travel time will be introduced to evaluate the driving performance of the trained policy at this intersection.
The passing rate is defined as the ratio of the successful pass time to the total 200 runs, wherein one successful pass means the automated vehicle can ride out of the intersection without collision with other participant, roads and the traffic lights violation. And the travel time is evaluated by the average time used to pass the intersection, starting from entering the intersection at the stop line.

The test results of the speeding case are shown in Table \ref{table.comp_speeding}. We can see that the driving policy of APG shows a stronger resistance to the overspeeding behavior of surrounding vehicles while that of DPG suffers a sharp drop on the passing rate. APG can handle totally the speeding by 10\% and the passing rate maintains steady when the speeding degree changes from 20\% to 50\%.
From the travel time, APG has learned to yield to the aggressive vehicles by decelerating and waiting them to pass firstly, which needs to sacrifice the passing efficiency.
This driving strategy provably seems reasonable because it can assure safety of the automated vehicle as much as possible.
\begin{table}[ht]
\centering
\caption{Test result of overspeed cases}
\begin{tabular}{l|l|c|c}
\hline
             &   Methods    &Passing rate   &Travel time (s)     \\ 
\hline
\multirow{2}*{Speeding by 10\%}  & APG  & 100\% & 13.08±2.35\\
                    & DPG       & 90\%         & 12.30±3.96   \\ 
\hline
\multirow{2}*{Speeding by 20\%}  & APG  & 90\%
 & 13.38±4.98   \\
                    & DPG       & 82\%
         & 12.08±1.71    \\ 
\hline
\multirow{2}*{Speeding by 50\%}  & APG  & 90\%
 & 16.18±4.85 \\
                    & APG       & 75\%
         & 10.33±3.28 \\ 
\hline
\end{tabular}
\label{table.comp_speeding}
\end{table}
In addition, the test results of rounding cases are shown in Table \ref{table.comp_turning}. We can conclude in this case the aggressive vehicles have greater influence on the ego because both policies demonstrate reduction with the increasing proportion of rounding vehicles. With the existence of adversarial training, the driving policy of APG has experienced the abnormal behavior more often during training, as the noise added on position, i.e., $\xi_x, \xi_y, \xi_\varphi$, might push the surrounding vehicle to round within the predictive horizon.
Therefore, it can still acquire a fair performance compared with the policy trained by DPG, which only possesses the 34\% passing rate when 50\% of vehicles attempt to behave without rules.
\begin{table}[ht]
\centering
\caption{Test result of rounding cases}
\begin{tabular}{l|l|c|c}
\hline
             &   Methods    &Passing rate   &Travel time (s)    \\ 
\hline
\multirow{2}*{10\% vehicles' rounding}  & APG  & 96\% & 12.83±3.07    \\
                    & DPG       & 80\%         & 12.71±2.88  \\
\hline
\multirow{2}*{20\% vehicles' rounding}  & APG  & 94\%
 & 13.38±3.91   \\
                    & DPG       & 72\%
         & 12.30±3.96  \\ 
\hline
\multirow{2}*{50\% vehicles' rounding}  & APG  & 84\% & 13.60±3.58 \\
                    & DPG       & 34\% & 10.50±4.94 \\ 
\hline
\end{tabular}
\label{table.comp_turning}
\end{table}

To sum up, we consider the uncertainty of the simulation model and introduce the adversarial training to enhance the driving performance, indicating that this will not only bring promotion at the training environment, but also access more resistances to disturbs from environment.

%% file: content/6Conclusion.tex
\section{Conclusion}
\label{sec:conclusion}
This paper focuses on the decision-making and control for signalized intersections, which are crucial and challenging for the popularity of autonomous driving.
To that end, we firstly design a general static path planner for the intersection scenarios, which consists of the route planning and velocity planning. It can generate smooth and trackable paths for the diversified intersections and feature high efficiency.
Secondly, we construct a constrained OCP for integrated decision and control wherein the bounded uncertainty of dynamic models is considered to capture the randomness of driving environment. APG is proposed to solve this OCP in which the adversarial policy is introduced to provide disturbances by seeking for the most severe uncertainty and the driving policy learns to handle this situation by competition.
Finally, a comprehensive system is established to conduct training and generalization test wherein the perception module is introduced and the human experience is incorporated to to solve the yellow light dilemma.
Results indicate that the trained policy can handle the yellow signal lights flexibly and realize smooth and efficient passing with a humanoid paradigm. Besides, the proposed APG enables the large-margin improvement of the resistance to the abnormal behavior of traffic participants and can ensure a high safety level for the autonomous driving.
About the future work, we will build large-scale intersections to make more powerful driving policy, and conduct real vehicle experiment to further verify its effectiveness.